\documentclass[journal]{IEEEtran}

\usepackage{graphicx}

\usepackage{cite}
\usepackage{pbox}
\usepackage{amsmath,amssymb,amsfonts}
\usepackage{algorithmic}
\usepackage{textcomp}
\usepackage{xcolor}
\usepackage{color}
\usepackage{multirow}
\usepackage{mathrsfs}
\usepackage{soul}

\usepackage[colorlinks,linkcolor=blue]{hyperref}

\soulregister\cite7 
\soulregister\uppercase7
\soulregister\expandafter7
\soulregister\romannumeral7

\ifCLASSINFOpdf

\else
 
\fi

\begin{document}

\title{Non-Homogeneous Haze Removal via Artificial Scene Prior and Bidimensional Graph Reasoning}

\author{Haoran Wei, Qingbo Wu,~\IEEEmembership{Member,~IEEE,}
        Hui Li,
        King Ngi Ngan,~\IEEEmembership{Fellow,~IEEE,}  \\
        Hongliang Li,~\IEEEmembership{Senior Member,~IEEE,}
        Fanman Meng,~\IEEEmembership{Member,~IEEE} and 
        Linfeng Xu
        % <-this % stops a space
\thanks{Haoran Wei and Qingbo Wu contributed equally to this work. The authors are with the School of Information and Communication Engineering, University of Electronic Science and Technology of China, Cheng Du, 611731, China (e-mail: hrwei@std.uestc.edu.cn; qbwu@uestc.edu.cn; huili@std.uestc.edu.cn;\\ knngan@ee.cuhk.edu.hk; hlli@uestc.edu.cn; fmmeng@uestc.edu.cn;\\lfxu@uestc.edu.cn). {(Corresponding authors: Qingbo Wu and Fanman Meng)}.}% <-this % stops a space
}

% \markboth{WEI \textit{\MakeLowercase{et al.}}: Non-Homogeneous Haze Removal via Artificial Scene Prior and Bidimensional Graph Reasoning}%
% {Shell \MakeLowercase{WEI \textit{et al.}}: Non-Homogeneous Haze Removal via Artificial Scene Prior and Bidimensional Graph Reasoning}

\maketitle

\begin{abstract}
Due to the lack of natural scene and haze prior information, it is greatly challenging to completely remove the haze from a single image without distorting its visual content. Fortunately, the real-world haze usually presents non-homogeneous distribution, which provides us with many valuable clues in partial well-preserved regions. In this paper, we propose a Non-Homogeneous Haze Removal Network (NHRN) via artificial scene prior and bidimensional graph reasoning. Firstly, we employ the gamma correction iteratively to simulate artificial multiple shots under different exposure conditions, whose haze degrees are different and enrich the underlying scene prior. Secondly, beyond utilizing the local neighboring relationship, we build a bidimensional graph reasoning module to conduct non-local filtering in the spatial and channel dimensions of feature maps, which models their long-range dependency and propagates the natural scene prior between the well-preserved nodes and the nodes contaminated by haze. To the best of our knowledge, this is the first exploration to remove non-homogeneous haze via the graph reasoning based framework. We evaluate our method on different benchmark datasets. The results demonstrate that our method achieves superior performance over many state-of-the-art algorithms for both the single image dehazing and hazy image understanding tasks. The source code of the proposed NHRN is available on https://github.com/whrws/NHRNet.
\end{abstract}

\begin{IEEEkeywords}
Non-Homogeneous haze removal, scene prior, gamma correction, artificial shot, bidimensional graph reasoning.
\end{IEEEkeywords}

\IEEEpeerreviewmaketitle

\section{Introduction}
Hazy weather easily causes poor image quality, which raises the risk of invalidating various outdoor computer vision applications including the object detection systems \cite{li2018end, chen2018domain, sindagi2020prior}, recognition systems \cite{wang2014joint} and so on. To overcome this issue, a lot of methods have been proposed to remove the haze from a single image.

\begin{figure}[t]
    \centering
    \includegraphics[width=0.47\textwidth]{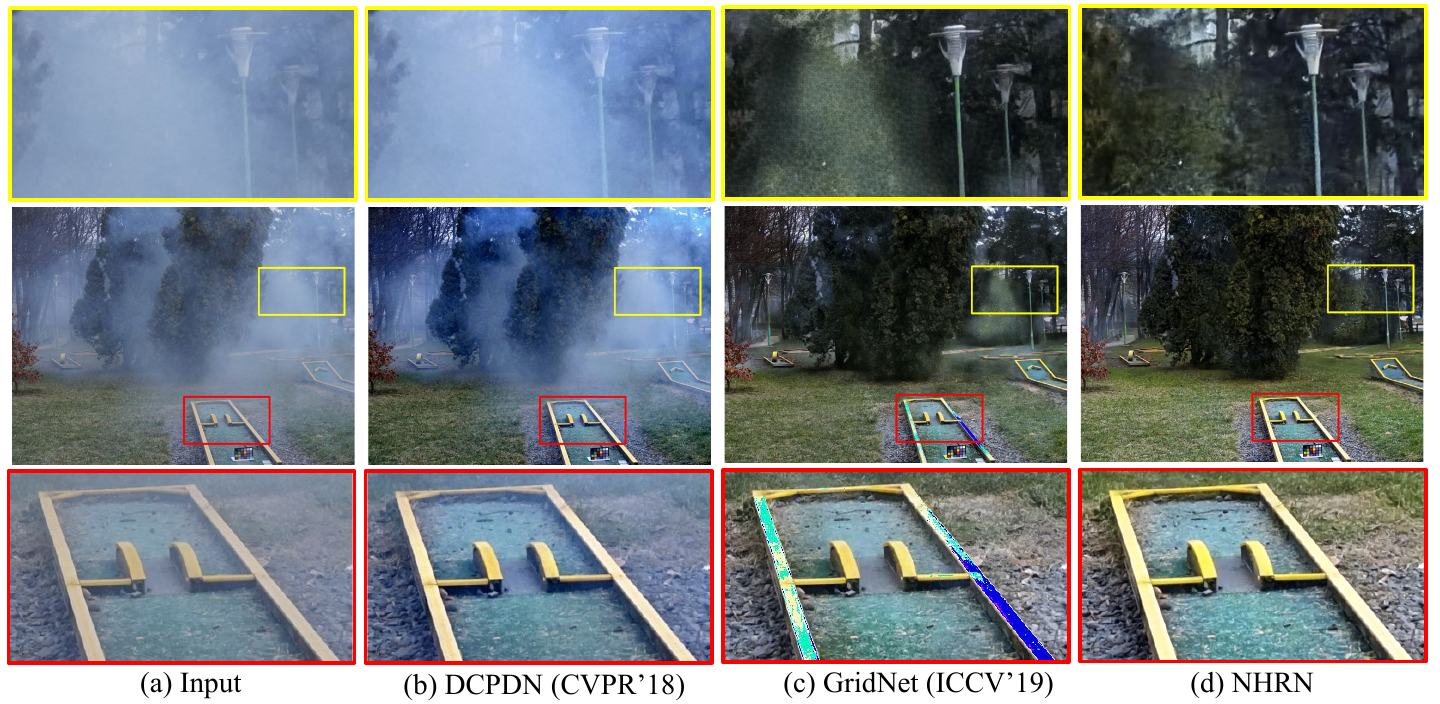}\caption{Haze removal results on a non-homgeneous real-world haze image from NTIRE 2020 dehazing challenge.}
    \label{fig:13}
\end{figure}

The early explorations are mainly developed from the atmospheric scattering model \cite{narasimhan2000chromatic, fattal2008single,narasimhan2002vision,tan2008visibility}, which formulates the haze-free image as an analytic solution with three variables, i.e., haze image, transmission map, and atmospheric light. Then, various prior assumptions are proposed to estimate the transmission map or atmospheric light, such as dark channel \cite{he2010single}, color attenuation \cite{zhu2015fast}, boundary constraint \cite{meng2013efficient} and so on. Limited by aforementioned specific assumptions, these prior based methods hardly perform well under diverse hazy scenes. 

Recently, with the rising up of deep learning, many data-driven methods have achieved impressive dehazing performance. Some methods follow the atmospheric scattering model, and derive more accurate transmission map and atmospheric light via designing convolution neural networks with BReLU \cite{Cai2016DehazeNet}, coarse-scale and fine-scale net \cite{ren2016single}, and densely connected pyramid structure \cite{zhang2018densely}. 

However, the estimation of the transmission map from a single hazy image is an ill-posed problem. Moreover, inaccurate estimation of the transmission map will interfere the atmospheric light estimation, which leads to produce undesired haze artifacts. To reduce the error accumulation, many methods proposed to directly regress the haze-free image. Such as Li \textit{et al}. \cite{li2017aod} applies a light-weight CNN, Ren \textit{et al}. \cite{ren2018gated} and Chen \textit{et al.} \cite{chen2019gated} perform a gated fusion module, Liu \textit{et al}. \cite{liu2019dual} adopts a dual residual connection structure, Liu \textit{et al}. \cite{liu2019griddehazenet} introduces multi-scale network with attention units, Dong \textit{et al}. \cite{dong2020fd} develops a fusion discriminator, and Dong \textit{et al}. \cite{dong2020multi} utilizes a dense feature fusion module.

Although different network architectures are explored, these deep learning based methods share two common issues, i.e., limited scene prior and contextual information utilization. As shown in Fig. 1, the scattering model based deep network DCPDN \cite{zhang2018densely} is unable to remove the haze, while fully end-to-end deep network GridNet \cite{liu2019griddehazenet} has limited ability to restore details and colors for non-homogeneous haze. Firstly, a single image is lack of depth and structure priors, which brings great difficulty to remove non-homogeneous haze with the atmospheric scattering model. Inspired by the success of multiple images based dehazing methods \cite{schechner2003polarization, schechner2001instant, shwartz2006blind, narasimhan2003contrast, galdran2018image}, we propose the artificial scene prior module to mimic multiple shots of the same scene captured under different exposure conditions, which do not require additional sensors like  \cite{schechner2003polarization, narasimhan2003contrast, nayar1999vision}. Furthermore, we generate artificial multiple shots by utilizing gamma corrections iteratively. In this way, the spatially-variant visibility of non-homogeneous haze is considered as inappropriate exposure, which could be well addressed by the multi-exposure fusion framework of human vision system \cite{ying2017bio}. As shown in Fig. 2, the proper exposure condition of each local region is varying, e.g., the tower could capture a well-perceived texture under the exposure conditions of $S_{3}$ and $S_{4}$, while the tree is clearer under the exposure conditions of $S_{2}$. Then, the image dehazing could be achieved by appropriately fusing our artificial multiple shots, which enriches the scene priors from the exposure conditions.

Second, existing methods mainly cope with non-homogeneous haze by selectively preserving the features of each local region with the guidance of transmission or attention maps. When some local regions are occluded by dense haze, they fail to explicitly utilize the information from the other distant regions, even they contain similar or complementary scene prior. Inspired by the classical non-local mean filter \cite{buades2005non}, we propose the bidimensional graph reasoning module (i.e., spatial graph reasoning module SGR and channel graph reasoning module CGR) to fully utilize the contextual information in removing the non-homogeneous haze. On the one hand, the proposed module simplifies the pixel-wise similarity of non-local mean filter to more compact node-wise similarity, which is computationally efficient. On the other hand, we extend the non-local information propagation from the spatial space to the channel space, which brings richer information compensation. As shown in Fig. 3, give an input image, in the spatial space, a collection of pixels in the dense haze region can interact with the long range clear pixels region, which focuses on the similar structures, e.g., the windows contaminated by haze and the long range well-perceived window. Similarly, in the channel space, these distant channels that responded to global content-related information (e.g., textures and edges) can interact with each other. To the best of our knowledge, this is the first exploration to remove non-homogeneous haze with this graph reasoning framework.

\begin{figure}[t]
    \centering
    \includegraphics[width=0.35\textwidth]{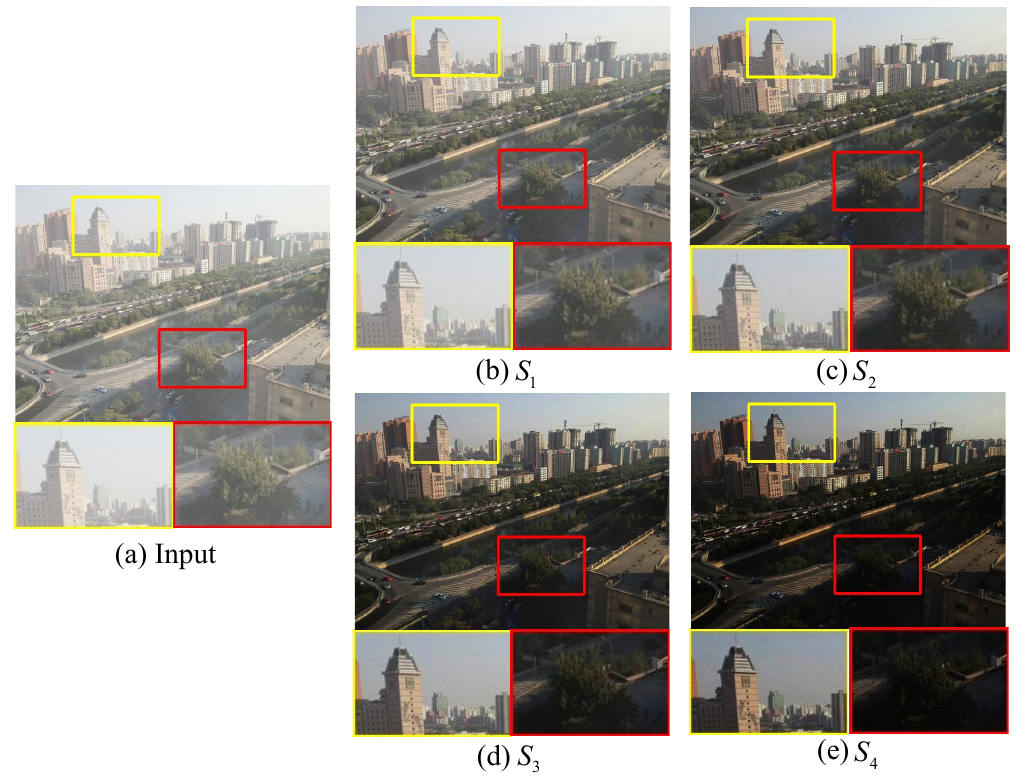}\caption{Visualization of the proposed artificial scene prior: ${S_{i}}|_{1\leq i\leq4}$ shows four different exposure versions.}
    \label{fig:18}
\end{figure}

\begin{figure}[t]
    \centering
    \includegraphics[width=0.48\textwidth]{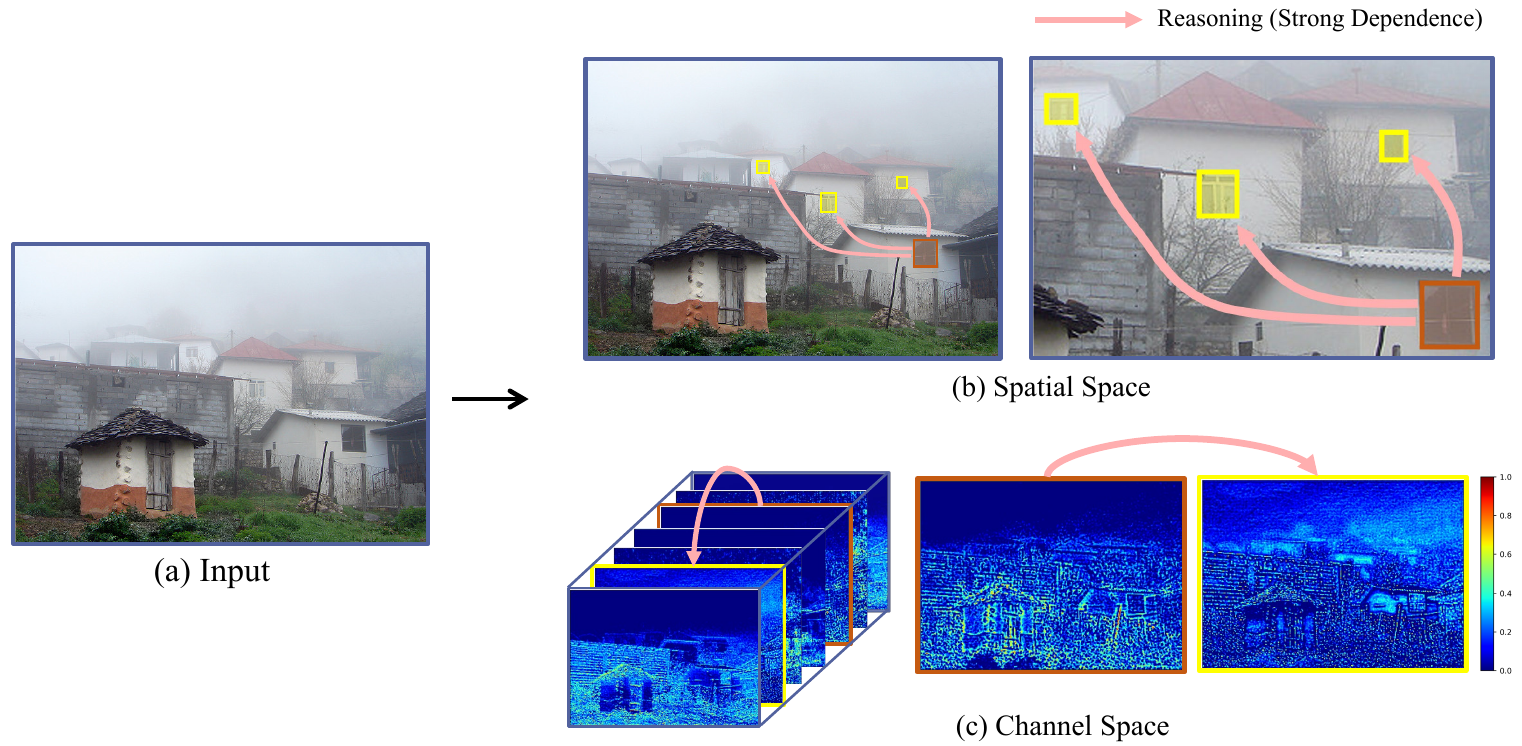}\caption{Visualization of the proposed bidimension reasoning, which aims to capture the long range dependencies among spatial and channel space.}
    \label{fig:14}
\end{figure}

In a nutshell, instead of relying on the atmospheric scattering model, we consider the single image dehazing as a joint artificial multi-exposure fusion and non-local filtering task, which is realized by the iterative gamma correction and graph reasoning framework. In comparison with our previous work in \cite{wei2020single}, the proposed Non-Homogeneous Haze Removal Network (NHRN) improves the dehazing performance from the following four aspects: 

$\bullet$ \textbf{Artificial Scene Prior}: We replace the exponential and logarithmic transformations with iterative gamma corrections, which is beneficial to reduce the parameter number and redundancy between different artificial shots. Meanwhile, in view of the potential noise, a gated fusion module is applied to selectively incorporate these artificial shots into the input features.

$\bullet$ \textbf{Spatial Projection Matrix}: For SGR module, we derive a dynamic spatial projection matrix via the similarity of multiple predefined anchors and pixel-wise features rather than a convolutional layer, which generates the node by aggregating the pixels with the same filter parameters regardless of the visual content change. In this way, we can adaptively enhance the consistency of all pixels within each node and reduce the risk of noise propagation when a node contains both the clear and contaminated pixels.

$\bullet$ \textbf{Unique Adjacency Matrix}: In \cite{wei2020single}, the graph reasoning is conducted via two 1$\times$1  convolutions, whose parameters become constant after training. It results that different images share the same adjacency matrix. In this paper, we derive the unique adjacency matrix for each image based on the pairwise similarity of its all nodes, which improves the adaptability of SGR and CGR in coping with diverse hazy scenes. 

$\bullet$ \textbf{Comprehensive Verification}: We conduct more extensive experiments on both synthetic and real-world hazy images, which verify the effectiveness of the proposed method using specifically designed image dehazing quality assessment metrics. Meanwhile, beyond perceptual quality evaluations, we conduct experiments to investigate the effect of dehazing methods in the detection and semantic segmentation tasks under a hazy scene.

\begin{figure*}[ht]
    \centering
    \includegraphics[width=0.8\textwidth]{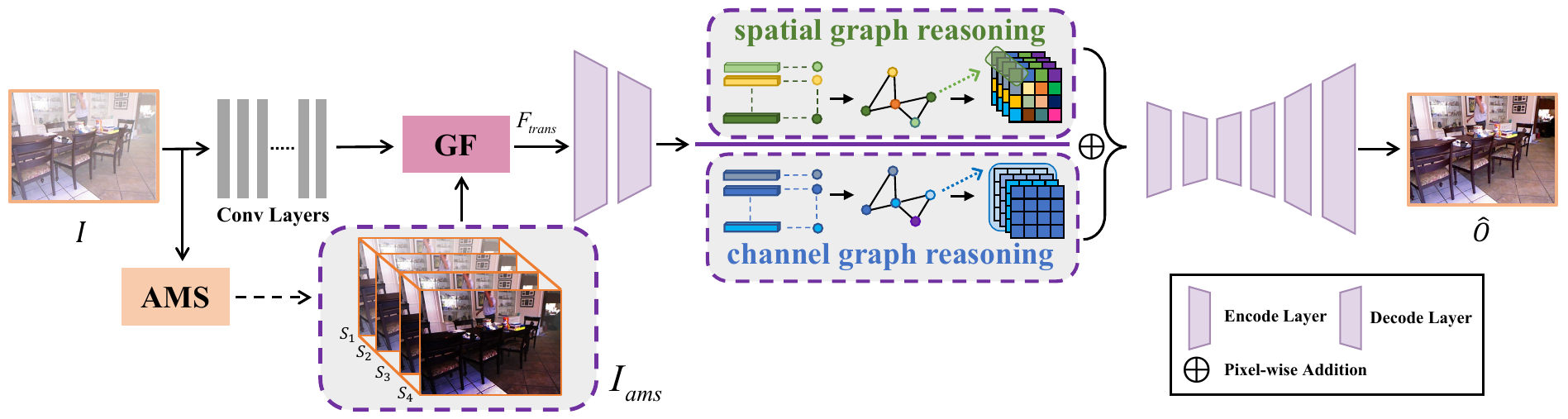}\caption{The overall scheme of our proposed network, it starts by generating artificial multiple shots using the AMS module (see Fig. 5). Next, we apply a gated fusion module (GF) to incorporate the obtained prior into the features of the input image. In the encoder layers, we adopt two context graph reasoning modules: the spatial graph reasoning module (SGR) and channel graph reasoning module (CGR) (see Fig. 6.). SGR and CGR aim to conduct non-local filtering in the spatial and channel dimensions of feature maps, which models their long-range dependency and propagates the natural scene prior between the well-preserved nodes and the nodes contaminated by haze.}
    \label{fig:15
    }
\end{figure*}

\section{Related Work}
\subsection{Single Image Dehazing}
Various prior methods have been proposed for image dehazing. DCP \cite{he2010single} is an effective strategy to restore the transmission map using the statistical prior of the lowest pixel values in the color channels. Meng \textit{et al}. \cite{meng2013efficient} estimates the transmission map by adding boundary constraint and contextual regularization. CAP \cite{zhu2015fast} applies the relationship between brightness and saturation to estimate the transmission map. IDE \cite{ju2021ide} introduces a light absorption coefficient to the atmospheric scattering model. Wu \textit{et al}. \cite{wu2019learning} proposes an interleaved cascade of shrinkage fields to reduce noise during recovering the transmission map and the scene radiance. {Ganguly \textit{et al}. \cite{ganguly2021single} presents a single image dehazing method by cascading a new atmospheric scattering model and a sparse haze model.} Further, learning-based methods are becoming popular, thanks to the strong nonlinear capabilities of deep networks. DehazeNet \cite{Cai2016DehazeNet}, MSCNN \cite{ren2016single} and DCPDN \cite{zhang2018densely} design CNNs to restore the haze-free images by using the atmospheric scattering model. Furthermore, AOD-Net \cite{li2017aod} applies a light-weight CNN to restore the clear images. GCANet\cite{chen2019gated} leverages a gated-fusion subnetwork to fuse the middle features extracted from different convolutional layers. DuRN \cite{liu2019dual} introduces a novel style of dual residual connection for image dehazing. GridNet \cite{liu2019griddehazenet} proposes the attention-based multi-scale dehazing network. {EDN \cite{yu2020ensemble} proposes a DenseNet based single-encoder four-decoders structure.} FD-GAN \cite{dong2020fd} develops a fully end-to-end generative adversarial network with a fusion discriminator for image dehazing. MSBDN \cite{dong2020multi} designs a dense feature fusion module, which remedies the missing spatial information from high-resolution features. RDN \cite{li2020deep} designs a deep retinex dehazing network, which jointly estimates the residual illumination map and the haze-free image. RATNet \cite{li2020region},\cite{9333643} proposes a region-adaptive two-shot network, which follows a coarse-to-fine framework. DM$^{2}$F-Net \cite{deng2019deep} presents a deep multi-model fusion network to achieve the image dehazing. Meanwhile, some methods propose to enrich the scene prior information of the input by various preprocessing operations. AMEF \cite{galdran2018image} applies a multi-scale Laplacian pyramid to fuse a series of gamma correction operations, which compensates valid information with each other. GFN \cite{ren2018gated} performs a gated network with fusing enhancement versions in three ways for preprocessing operations, including white balanced, contrast enhanced and gamma corrected. However, these methods ignore that real-world haze usually exhibits non-homogeneous distribution, so that well-preserved information could provide us with many valuable clues. However, convolution operations suffer from the limitation, which fails to explicitly utilize the information from the other distant regions, even they contain important scene prior.

\subsection{Graph-Based Method}
Recently, Graph-based methods have been very popular and aim to model long-range dependencies through graph-structured data. In particular, an image could be regarded as a regular graph-structure data. CRFs \cite{chandra2017dense} is an effective graph model used as post-processing for image segmentation. Besides, Graph Convolutional Networks (GCN) exhibits superior performance on the task of reasoning relation. Specifically, Kipf \textit{et al.} \cite{kipf2016semi} encodes the graph structure to reason relation between graph nodes for semi-supervised classification. Wang \textit{et al.} \cite{wang2018videos} exploits GCN captures relations between objects in video recognition tasks. Xu \textit{et al.} \cite{xu2019spatial} proposes to use GCN for large-scale object detection, which discovers and incorporates key semantic and spatial relationships for reasoning over each object. Furthermore, Chen \textit{et al.} \cite{chen2019graph} adopts the reasoning power of graph convolutions to reason disjoint and distant regions without extra annotations for semantic segmentation.

\section{Proposed Method}

\subsection{Overview}

{The proposed network architecture is illustrated in Fig. 4, which consists of Artificial Scene Prior and Bidimensional Graph Reasoning.}

\subsubsection{Artificial Scene Prior} The hazy image \textit{I} is firstly preprocessed via the artificial multiple shots module (AMS), which generates artificial multiple shots and connects them along the channel dimension to obtain a set of exposed images ${I}_{ams}$, i.e.,
\begin{equation}
{I_{ams} = AMS(I) = \lbrack S_{1}\dots \circ  S_{i} \dots \circ  S_{m} \rbrack,}
\end{equation}
where $m$ is the total number of artificial shots, ${S_{i}}|_{1\leq i\leq m}$ is the $i$-$th$ level artificial shot and $\circ$ denotes the concatenation operation. Moreover, we apply the gated fusion module ($GF$) to fuse scene prior features from these exposed images ${I}_{ams}$ and the input $I$. We acquire the transformed features ${F}_{trans}$, which preserves the regions with important recovery information and filters extra noise, i.e.,
\begin{equation}
{F_{trans} =  GF({I}_{ams}, {I}).}
\end{equation}
{We further feed the transformed features ${F}_{trans}$ into the encoder convolutional layers $\mathscr{E}(\cdot)$ to obtain the encoding features $X$, i.e.,}
\begin{equation}
{X = \mathscr{E}({F}_{trans}).}
\end{equation}

\subsubsection{Bidimensional Graph Reasoning}
{We employ the spatial graph reasoning module (SGR) and channel graph reasoning module (CGR) to conduct non-local filtering over regions and over channels respectively, which models long-range dependency and propagates the natural scene prior between the well-preserved nodes and the nodes contaminated by haze.} 

{Specifically, we project $\textit{X}$ into the spatial or channel graph structure $\mathcal{G}{_{i}}=(\mathcal{V}{_{i}}, \mathcal{E}{_{i}}, {A}_{i})|_{i\in\{s,c\}},$ where the subscript $i$ indicates the graph reasoning used in different dimensional, i.e., $s$ indicates the spatial dimensional, and $c$ indicates the channel dimensional. $\mathcal{V}{_{i}}|_{i\in\{s,c\}}$ denotes  its nodes, and $\mathcal{E}{_{i}}|_{i\in\{s,c\}}$ denotes related edges. Moreover, ${A}{_{i}}|_{i\in\{s,c\}}$ describes corresponding edge weights for the graph. The bidimensional graph reasoning based framework consists of three operations:} 

\textbf{Graph Projection}: {we aim to project $X$ into a set of spatial or channel graph nodes, respectively, i.e.,}
\begin{equation}
{V_{i} = \mathcal{G}_{proj}(X) = B_{i} \times X,}
\end{equation}
{where $\mathcal{G}_{proj}$ denotes the graph projection, $\times$ denotes the matrix multiplication, ${B_{i}}|_{i\in\{s,c\}}$ denotes the corresponding graph projection matrix, and ${V_{i}}|_{i\in\{s,c\}}$ denotes node features for the corresponding graph ${\mathcal{G}_{i}}$. Based on node features ${V_{i}}$, we construct the edges ${\mathcal{E}_{i}}$ between each pair of the nodes by computing the adjacency matrix.} {Taking into account the difference in dependency between the nodes, we choose to construct the directed graph, where the matrix element  ${a}{_{i}}(p,q)$ of ${A}{_{i}}$ represents the connectivity from $p$-$th$ node features $v_{i}(p)$ to the $q$-$th$ node features $v_{i}(q)$, i.e.,}
\begin{equation}
{{a}_{i}(p,q) = \frac{exp[\theta(v_i(p)) \theta^{'}(v_i(q))^\mathsf{T}]}{\sum_{q=1}^{N}exp[\theta(v_{i}(p)) \theta^{'}(v_{i}(q))^\mathsf{T}]},}
\end{equation}
{where ${A_{i}}=\{{a_{i}}(p,q)\}$, $p = 1,2 \dots, N$, $q = 1,2 \dots, N$, $N$ denotes the number of graph nodes, $\theta(\cdot)$ and $\theta^{'}(\cdot)$ denote two linear embeddings of different parameters.}

\textbf{Graph Reason}: {We learn the connectivity between nodes from ${V_{i}}$, i.e., the relations over regions and over channels. Meanwhile, we reason over the relations by propagating information across nodes to model long-range dependency. We feed the node features ${V_{i}}$ into a graph convolution $\mathcal{G}_{gcn}$, i.e.,}
\begin{equation}
{\hat{V}_{i}= \mathcal{G}_{gcn}(V_{i}) = \sigma({A}_{i} \times V_{i} \times W_{i}),}
\end{equation}
{where $\sigma(\cdot)$ denotes the activation function ReLU, ${W_{i}}|_{i\in\{s,c\}}$ denotes the weights of the graph convolution, and ${\hat{V}_{i}}$ denotes the features acquired by the node-wise interaction. Since $\mathcal{G}$ has a receptive field of all nodes on the graph, and thus is able to capture the global context of the input in the reasoning.}

\textbf{Graph Reprojection}: {We reproject the spatial or channel node-wise interaction features in the graph space to the original pixel space, i.e.,}
\begin{equation}
{Y_{i} = \mathcal{G}_{reproj}(\hat{V}_{i}) = D_{i} \times \hat{V}_{i},}
\end{equation}
{where $\mathcal{G}_{reproj}$ denotes the graph reprojection, ${D_{i}}|_{i\in\{s,c\}}$ represents the corresponding graph projection matrix, and ${Y_{i}}|_{i\in\{s,c\}}$ represents the refined pixel-wise features.} {Therefore, our SGR and CGR can be expressed as}
\begin{equation} 
\label{eqn2}
  \begin{split}
 {Y_{i}}&{= \mathcal{G}_{reproj}(\mathcal{G}_{gcn}(\mathcal{G}_{proj}(X)))}\\
    &{= D_{i} \times \sigma[{A}_{i} \times (B_{i} \times X) \times W_{i}].}
  \end{split}
\end{equation}
{In order to facilitate be incorporated into various backbone, we added a residual loop \cite{he2016deep}, i.e.,}
\begin{equation}
{F_{i} = Y_{i} + X,}
\end{equation}
{where $F_{i}$ represents the final pixel-wise features by reasoning.} {In the following, we feed the final pixel-wise features by spatial reasoning ${F}_{s}$ and the final pixel-wise features by channel reasoning ${F}_{c}$ to the remaining encoder layers and a decoder network $\mathscr{D}(\cdot)$, which generates the final dehazing image $\hat{O}$, i.e.,}
\begin{equation}
{\hat{O}=\mathscr{D}({F}_{s}+{F}_{c}).}
\end{equation}

\begin{figure}[t]
    \centering
    \includegraphics[width=0.47\textwidth]{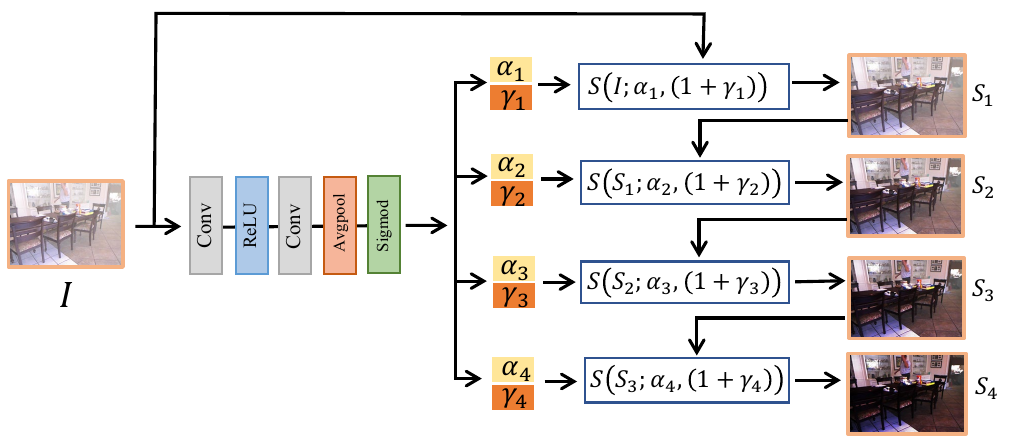}\caption{The schematic illustration about how to produce artificial multiple shots (denoted as $S_{1}, S_{2}, S_{3}$ and $S_{4}$) using the AMS module. Moreover, $S_{i+1} = S(S_{i};\alpha_{i+1},(1+\gamma_{i+1}))$ denotes the $(i+1)$-$th$ level artificial shots.}
    \label{fig:15}
\end{figure}

\subsection{Artificial Scene Prior}
{Inspired by the success of multiple images based dehazing methods \cite{schechner2003polarization, schechner2001instant, shwartz2006blind, narasimhan2003contrast, galdran2018image}, we use AMS module to iteratively produce a set of different exposure images in the same scene. Meanwhile, we apply the GF module to fuse the features from input and these exposure images, which aims to capture the underlying scene prior.}
\subsubsection{Artificial Multiple Shots}
The AMS module is a learnable preprocessing operation, which compensates for their high-frequency components in multiple degrees. Inspired by the traditional image enhancement process, we attempt to obtain multiple high-frequency compensation versions using gamma correction for the input hazy image. Meanwhile, the parameters of high-frequency compensation are adaptively determined by the input image.

In view of the diversity of the haze distribution, our method iteratively generates artificial shots in different levels, which simulates the images captured under different exposure conditions, i.e.,
\begin{equation}
{S_{i+1} = S_{i} + \alpha_{i+1}(S_{i}^{1+\gamma_{i+1}} - S_{i}),}
\end{equation}
where $S_{i}$ represents the $i$-$th$ level artificial shot, $S_{0}=I$, and $\alpha$ and $\gamma$ $\in[0,1]$ are the trainable enhancement parameters, which adjust the magnitude and control the high-frequency compensation level. As shown in Fig. 5, to learn the mapping between an input image and its suitable multiple sets of high-frequency parameters $\alpha$ and $\gamma$, we feed the input image to two $3 \times 3$ kernel convolutions and the ReLU activation function followed by an average pooling layer and a Sigmod function. Finally, we combine multiple artificial shots to obtain $I_{ams}$, which adapts to different enhancement requirements.

\subsubsection{Gated Fusion module}
We regard these artificial multiple shots as the underlying scene prior of the hazy image. Since each $\alpha$ and $\gamma$ are used for all pixels, it is a global adaptive high-frequency compensation. Notably, the generated artificial multiple shots may introduce noise interference for some local regions while using global mapping. {As shown in Fig. 2, the tree is clear when under the artificial shots $S_{2}$. However, the visual performance of the tree in other artificial shots are brighter or darker.}

To address this problem, we apply the GF module to preserve pivotal prior for better dehazing optimization. Firstly, GF applies the convolution operation to obtain $F_{ams}$ and $F_{haze}$ by extracting features from these artificial multiple shots $I_{ams}$ and the original input image $I$, respectively.

\begin{figure}[t]
    \centering
    \includegraphics[width=0.47\textwidth]{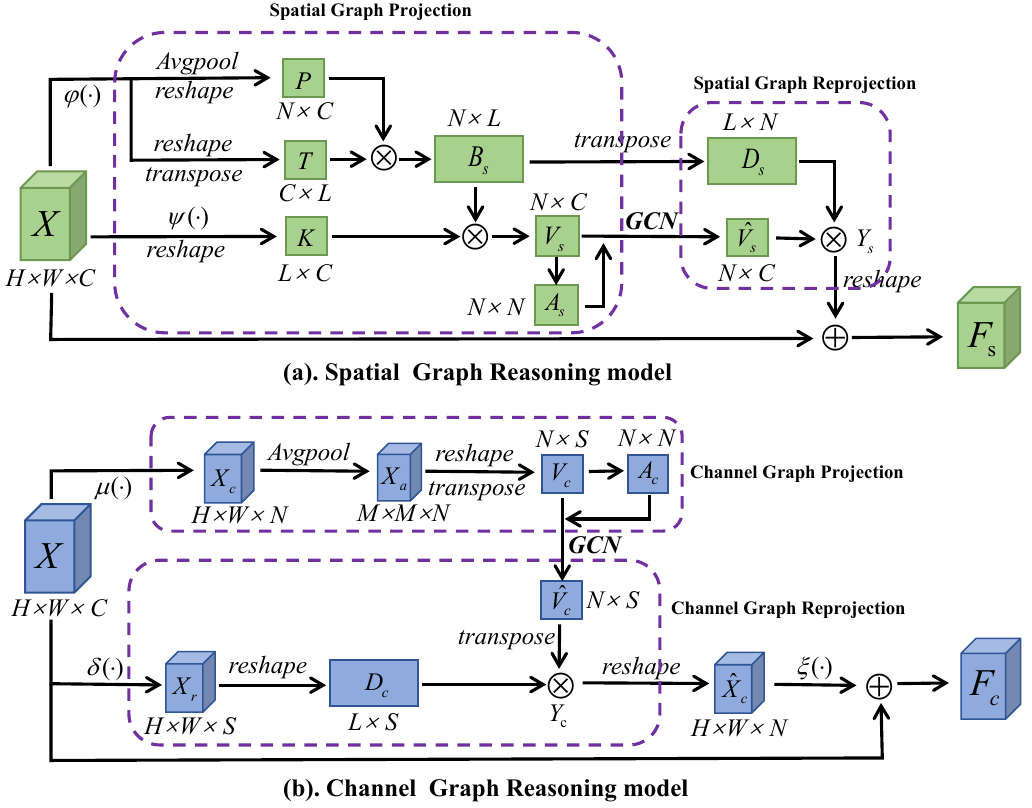}\caption{The details of Spatial Graph Reasoning module and Channel Graph Reasoning module are illustrated in (a) and (b).}
    \label{fig:16}
\end{figure}

Second, GF adopts a gated fusion operation to incorporate the features of artificial multiple shots $F_{ams}$ into the features of the hazy input $F_{haze}$. The outputs of the gated fusion operating are two different importance weights ($\mathcal{W}_{ams}, \mathcal{W}_{haze}$), which correspond to each feature level, respectively.
\begin{equation}
{\mathcal{W}_{ams}, \mathcal{W}_{haze} = \mathcal{W} (F_{ams} \circ F_{haze}),}
\end{equation}
\begin{equation}
{F_{trans} = \mathcal{W}_{ams} * {F}_{ams} + \mathcal{W}_{hazy}* {F}_{haze},}
\end{equation}
where $\mathcal{W}(\cdot)$ represents the gated operation, which consists of two convolutional layers with kernel size 3x3 and a Sigmod function. Meanwhile, the channel dimension of $\mathcal{W}_{ams}$ and $\mathcal{W}_{haze}$ is one, respectively.

\subsection{Bidimensional Graph Reasoning}
We employ the pretrained ResNet-50 \cite{he2016deep} as the encoder network. The decoder consists of five consecutive deconvolutional layers \cite{zeiler2011adaptive} to restore the original resolution. Considering that haze distribution is non-homogeneous across spatial location, we aim to build the long-range interactions between different regions with similar structure. Meanwhile, it is critical to capture the relationship between channels. Inspired by the graph convolutional network, we develop the Spatial Graph Reasoning module and Channel Graph Reasoning module to model the long-range dependency 
over regions and over channels on a graph, respectively. 

\subsubsection{Spatial Graph Reasoning module}
As illustrated in Fig. 6 (a), we exploit the correlation among distinct regions via Spatial Graph Projection, Spatial Graph Reasoning, and Spatial Graph Reprojection.

\textbf{Spatial Graph Projection}: For an input feature tensor $\textit{X}\in\mathbb{R}^{H \times W \times C}$, we aim to use $B_{s}$ to project $\textit{X}$ into a set of spatial nodes in a graph. Specifically, we first adopt a $1\times1$ convolution $\psi(\cdot)$ as a linear embedding, resulting in $K = \psi(X)$. We then reshape $K$ to $\in\mathbb{R}^{L \times C}$. Next, we utilize $K$ to replace $X$ in the $Eq. (4)$. Specifically, we multiply $B_{s}$ and $K$ to obtain spatial node features $V_{s} \in\mathbb{R}^{N \times C}$ in the spatial graph, where $N$ is the node numbers, and each node can be represented by a $C$-dimensional vector.
\begin{figure}[t]
    \centering
    \includegraphics[width=0.47\textwidth]{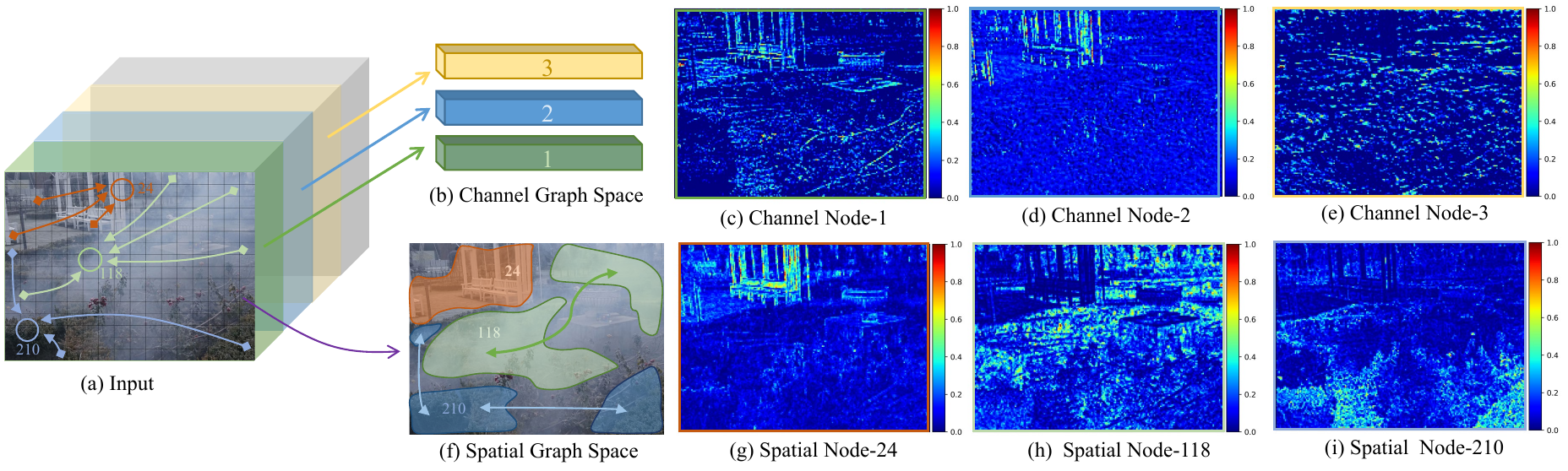}\caption{Illustration of the projection process of forming spatial and channel graph nodes. Note that the corresponding spatial anchor position is marked on (a).}
    \label{fig:17}
\end{figure}

Considering the pixel features consistency within each node, we derive a dynamic $B_{s}$, which could group pixels into coherent regions. In particular, we capture the similarity of pixel-wise features to guide the allocation for each pixel. However, the complexity of computing is large with the increasing number of pixels $H \times W$. To address this issue, we uniformly divide the spatial feature map into $N$ grids as shown in Fig. 7 (a), and use the average pooling of each grid to represent its anchor. In addition to reducing computational complexity, the average pooling could promote the compact representation over local features to reduce redundancy. Then, the similarity is measured between the anchors and the pixel-wise features. Specifically, to generate a dynamic projection matrix $B_{s}$, we first apply a $1\times1$ convolution $\varphi(\cdot)$ as a linear embedding to obtain $\varphi(X)$. Second, we build $N$ anchors of nodes by an average pooling operation $\mathcal{P}(\cdot)$, which reduces the spatial dimensions of $\varphi(X)$ from ($H \times W$) to ($Q \times Q$) of $\mathcal{P}(\varphi(X))\in\mathbb{R}^{Q \times Q \times C}$. We then flatten $\mathcal{P}(\varphi(X))$ to obtain the $N$ anchor features $P\in\mathbb{R}^{N \times C}$, $N = Q \times Q$. Meanwhile, we reshape and transpose $\varphi(X)$ to obtain the pixel-wise features $T$ $\in\mathbb{R}^{C \times L}$, $L = H\times W$. Finally, we derive $B_{s}$ via the similarity of multiple predefined anchors $P$ and pixel-wise features $T$. In particular, we take the multiplication of $P$ and $T$ to obtain $B_{s}$. Additionally, we choose the Softmax function for normalization in the node direction, i.e.,
\begin{equation}
{B_{s} = \frac{exp(P \times T)}{\sum_{n=1}^{N}exp(P \times T)},}
\end{equation}
where $n=1,2...,N$ denotes the $n$-$th$ anchor, $\times$ denotes the matrix multiplication, and $B_{s}\in\mathbb{R}^{N \times L}$. Fig. 7 (f) shows the projection process of forming spatial graph nodes. Fig. 7 (g), (h) and (i) are the weights for these projection maps of $B_{s}$ (i.e., Node-24, 118 and 210). Meanwhile, Node-24 aggregates pixels with similar features as Anchor-24, which focuses on the pavilion structure in clear regions. Node-118 and 210 aggregate grass structures from dense haze and clear regions respectively, which are similar to Anchor-118 and 210. In other word, $B_{s}$ forms each node by aggregating pixels with similar features. In particular, each node can represent an arbitrary region with the similar pixel-wise features. Finally, we use $Eq. (4)$ to obtain the $N$ node features $V_{s} \in\mathbb{R}^{N \times C}$, with each node represented by a $C$-dimensional vector. Moreover, we learn the directed adjacency matrix ${A}_{s}\in\mathbb{R}^{N \times N}$ by $Eq. (5)$.

\textbf{Spatial Graph Reasoning}:
We employ $Eq. (6)$ to update the feature information for each node, which is obtained from other nodes. Specifically, $W_{s}\in\mathbb{R}^{C\times C}$ is the weight matrix, and $\hat{V}_{s}\in\mathbb{R}^{N\times C}$ denotes the updated node features after diffusing information across nodes.

\textbf{Spatial Graph Reprojection}:
In the following, we perform the features reprojection from graph space to pixel space via spatial graph reprojection. For simplicity, we reuse the spatial projection matrix $B_{s}$. In particular, we obtain the spatial reprojection matrix $D_{s}$ by transposing $B_{s}$. We further project the graph node features $\hat{V}_{s}\in\mathbb{R}^{N \times C}$ to the pixel-wise features $Y_{s}\in\mathbb{R}^{L \times C}$ by the spatial projection matrix $D_{s}$ by $Eq. (7)$. We then reshape the size of $Y_{s}$ to $\in\mathbb{R}^{H \times W \times C}$ to get the residual value for $X$. Finally, we add the input $X$ and the residual to obtain ${F}_{s}$.

\subsubsection{Channel Graph Reasoning module}
Similar to SGR, we exploit the correlation among distinct channels via Channel Graph Projection, Channel Graph Reasoning, and Channel Graph Reprojection. The structure of CGR is illustrated in Fig.6 (b).

\textbf{Channel Graph Projection}: We perform the features projection on the channel dimension. In particular, we project each channel feature into a single channel graph node. We first feed $X$ into a $1 \times 1$ convolution layer $\mu(\cdot)$ to generate $\textit X_{c}\in\mathbb{R}^{H \times W \times N}$, where $N$ determines the number of nodes. Considering that $L = H\times W$ is too large, we utilize an average pooling $\mathcal{P}(\cdot)$ to process the features $X_{c}$, which reduces the scale for $H$ and $W$, i.e., $X_{a} = \mathcal{P}(X_{c})$. $\textit X_{a}\in\mathbb{R}^{M \times M \times N}$, $M$ denotes the output scale after the pooling operation. Since the number of channels is limited, we directly project each channel feature into a node feature by $B_{c}$, so that the limited number of nodes will not increase a lot of calculation in the subsequent reasoning. Moreover, different from $B_{s}$, $B_{c}$ can be expressed as a simple feature size transformation to obtain node-wise features $\textit V_{c}$. Specifically, we reshape $\textit X_{a}$ to $\in\mathbb{R}^{S \times N}$ ($S = M\times M$) and obtain the transpose of it, i.e., $\textit V_{c}\in\mathbb{R}^{N \times S}$. $\textit V_{c}$ represents the $N$ node features of the input $\textit X$ projected in the channel dimension, with each node represented by a $S$-dimensional vector. Fig. 7 (b) shows the projection process of forming channel nodes. And Fig. 7 (c),(d) and (e) are the feature maps before pooling of channel Node-1, 2 and 3. Similarly, we exploit $Eq. (5)$ to generate a directed adjacency matrix ${A}_{c}\in\mathbb{R}^{N \times N}$ in the graph. 

\begin{table*}[!h]
	\vspace{-0.2cm}
	\centering
	\caption{Quantitative comparisons on real-world and synthetic dehazing datasets. It demonstrates our method outperforms previous dehazing methods in term of PSNR, SSIM and LPIPS, where $\uparrow$ means the higher the better, and $\downarrow$ means the lower the better.}
	\resizebox{0.65\linewidth}{!}{
		\begin{tabular}{|c|c|c|c|c|c|c|c|c|c|c|c|c|c|c|c|}
			\hline
			& \multicolumn{3}{|c|}{SOTS indoor} & \multicolumn{3}{|c|}{SOTS outdoor} & \multicolumn{3}{|c|}{TestA-DCPDN} & \multicolumn{3}{|c|}{NH-HAZE}&\multicolumn{3}{|c|}{O-HAZE} \\
			\hline
			Method & PSNR $\uparrow$ & SSIM $\uparrow$& LPIPS $\downarrow$ & PSNR$\uparrow$ & SSIM $\uparrow$ & LPIPS $\downarrow$ & PSNR$\uparrow$ & SSIM $\uparrow$ & LPIPS $\downarrow$ & PSNR $\uparrow$& SSIM $\uparrow$& LPIPS $\downarrow$& PSNR $\uparrow$& SSIM $\uparrow$& LPIPS $\downarrow$  \\
			\hline
			\hline
			DCP \cite{he2010single}           & 16.62 & 0.818 & 0.099 & 21.15 & 0.896 & 0.105 & 13.91 & 0.864 & 0.142 & 12.60 & 0.471& 0.465 & 14.30 & 0.587 & 0.438\\
			CAP \cite{zhu2015fast}          & 19.05 & 0.840 & 0.102 & 22.43 & 0.920 &0.058  & 15.86 & 0.726 & 0.186 & 13.01 & 0.445 & 0.556 & 17.46 & 0.660 & 0.391 \\
			Meng \cite{meng2013efficient}         & 23.49 & 0.936 & 0.136 &15.51  & 0.797 & 0.186 & 24.33 & 0.904 & 0.172 & 13.12 & 0.492 & 0.499 &14.43&0.594 &0.495\\
			AMEF \cite{galdran2018image}          & 17.52 & 0.797 & 0.143 & 17.81 & 0.820 &0.143  & 18.27 & 0.812 & 0.154 & 13.82 & 0.506 & 0.509 &17.30&0.632&0.389\\
			\hline
			\hline
			DehazeNet \cite{Cai2016DehazeNet}     & 21.14 & 0.850 & 0.071 & 22.46 & 0.851 & 0.075 & 19.92 & 0.858 & 0.096 & 11.76 & 0.412 & 0.563&16.21&0.666&0.386 \\
			AOD-Net \cite{li2017aod}       & 20.86 & 0.879 & 0.303 & 23.36 & 0.917 & 0.074 & 20.46 & 0.838 & 0.183 & 10.97 & 0.369 & 0.654 &19.59&0.679&0.543\\
			GFN \cite{ren2018gated}           & 22.30 & 0.880 & 0.065 & 21.55 & 0.844 & 0.105 & 25.59 & 0.939 & 0.091 & 12.60 & 0.388 & 0.722 &22.58&0.737&0.393\\
			DCPDN \cite{zhang2018densely}        & 28.13 & 0.959 & 0.129 & 21.59 & 0.880 & 0.073 & 29.27 & 0.953 & 0.080 & 11.32 & 0.467 & 0.538&22.78&0.742& 0.375\\
			GCANet \cite{chen2019gated}         & 30.06 & 0.960 & 0.023 & 28.45 & 0.955 & 0.053 & 28.18 & 0.972 & 0.040 & 14.01 & 0.490 & 0.491 & 24.10 & 0.758 & \textcolor{red}{0.272}\\
			DuRN  \cite{liu2019dual}         & 32.12 & 0.980 & 0.016 & {29.22} & {0.970} & \textcolor{blue}{0.015} & 32.60 &\textcolor{blue}{0.983} & 0.045 & {19.38} & {0.681} &{0.318}&23.23&0.753&0.315\\
			GridNet \cite{liu2019griddehazenet}  & {32.16} & \textcolor{blue}{0.983} & {0.012} & 28.05 & 0.971 & 0.017 & {32.71} & 0.982 &{0.028} & 17.76 & 0.644 & 0.327&23.67 & 0.746 & 0.300\\
			EDN \cite{yu2020ensemble}  & {30.24} & {0.967} & {0.021} & 29.07 & 0.967 & 0.017 & {31.99} & 0.967 &\textcolor{red}{0.020} & 18.58 & 0.630 & 0.303 & 24.21 & 0.753 & \textcolor{blue}{0.274}\\
			FD-GAN \cite{dong2020fd}  & {22.14} & {0.894} & {0.082} & 23.35 & 0.924 & 0.075 & {20.88} & 0.884 &{0.121} & 14.56 & 0.500 & 0.480&16.34&0.574&0.428\\
			MSBDN \cite{dong2020multi}  & {32.80} & {0.980} & {0.014} & 27.27 & 0.957 & 0.022 & {28.89} & 0.927 &{0.044} & 19.18 & 0.669 & 0.307&23.97&0.760&0.306\\
			\hline
			\hline
			\textbf{MCN}\cite{wei2020single} & \textcolor{blue}{35.23} & \textcolor{red}{0.989} &  \textcolor{blue}{0.009}& \textcolor{blue}{30.46} & \textcolor{blue}{0.977} &\textcolor{blue}{0.015} & \textcolor{blue}{34.04} & \textcolor{red}{0.987} &{0.025} & \textcolor{blue}{19.78} & \textcolor{blue}{0.699} & \textcolor{blue}{0.299}&\textcolor{blue}{24.87}&\textcolor{blue}{0.782}&0.336\\
			\textbf{NHRN} & \textcolor{red}{35.60} & \textcolor{red}{0.989} &  \textcolor{red}{0.008}& \textcolor{red}{30.62} & \textcolor{red}{0.978} &\textcolor{red}{0.014} & \textcolor{red}{34.36} & \textcolor{red}{0.987} &\textcolor{blue}{0.023} & \textcolor{red}{20.37} & \textcolor{red}{0.707} & \textcolor{red}{0.286}& \textcolor{red}{25.13} & \textcolor{red}{0.791} & 0.328\\
			\hline
	\end{tabular}}
	\vspace{0.2cm}
	\label{result5}
\end{table*}

\textbf{Channel Graph Reasoning}: 
We exploit $Eq. (6)$ to reason the graph relationship to acquire the new node features $\hat{V_{c}} \in\mathbb{R}^{N \times S}$. $W_{c}\in\mathbb{R}^{S\times S}$ is the weight matrix, and $\hat{V}_{c}\in\mathbb{R}^{N \times S}$ denotes the updated node features after diffusing information across channel nodes.

\textbf{Channel Graph Repojection}: 
We perform the features reprojection from graph space to pixel space via channel graph reprojection. Considering that the operation of channel projecting to the nodes could cause some information to be lost, so we relearn a mapping function, which transforms the node features $\hat{V}_{c}$ to $\in\mathbb{R}^{L \times N}$. In particular, we generate $X_{r}$ from $X$ by a linear projection $\delta(\cdot)$, i.e., $X_{r} = \delta(X)$. Then we reshape and transpose $X_{r}$ as the channel reprojection matrix $D_{c}\in\mathbb{R}^{L \times S}$. Furthermore, we reproject the transpose of channel node features $\hat{V}_{c}$ to obtain pixel-wise features $Y_{c}$ by $Eq. (7)$, where $Y_{c}$ $\in\mathbb{R}^{L\times N}$ is the pixel-wise features. Furthermore, we reshape it to obtain $\hat{X}_{c}\in\mathbb{R}^{H \times W \times N}$. Additionally, we use another a $1\times1$ convolution layer $\xi(\cdot)$ to obtain the residual value from $\hat{X}_{c}$, which changes $N$ to $C$ to match the input dimension for $X$. Finally, we add the input $X$ and the residual to obtain ${F}_{c}$.

\section{Loss Function}
The final dehazing result is $\hat{O}$. $O$ denotes the clear image, and $Num$ denotes the number of pixels in an image. The image content loss ${L}_{c}$ and ssim loss ${L}_{ssim}$ are incorporated for training the proposed network, i.e.,
\begin{equation}
{L_c=\frac{1}{Num}||\hat{O}- {O}||^{2},}
\end{equation}
\begin{equation}
{L_{ssim}=1- SSIM(\hat{O}, O).}
\end{equation}
Hence, the total loss ${L}_{total}$ can be expressed as:
\begin{equation}
{L_{total}=L_c +\lambda L_{ssim},}
\end{equation}
where $\lambda$ is a parameter used to balance two loss functions and is set to 0.1 by default.

\section{Experimental Results}
\label{sec:pagestyle}
In this part, we first describe the benchmark databases, which contain both the synthetic and real-world haze images. Second, we expound the implementation details of our method. Next, we conduct quantitative and qualitative analysis to compare our dehazing network against state-of-the-art methods.  

\subsection{Training and Testing Datasets}
We evaluate the proposed method on publicly available benchmarks, including synthetic indoor and outdoor datasets RESIDE \cite{li2018benchmarking}, DCPDN-data \cite{zhang2018densely}, Foggy Cityscapes-DBF \cite{sakaridis2018model} followed by the realistic outdoor dataset O-HAZE \cite{ancuti2018haze}, realistic non-homogeneous dataset NH-HAZE \cite{ancuti2020nh} and DHQ \cite{min2019quality}.

The RESIDE dataset consists of paired synthetic indoor and outdoor hazy images. Specifically, we choose the RESIDE ITS and OTS-$\beta$ \cite{li2018benchmarking} as the train sets, respectively. 13990 indoor hazy images of ITS \cite{li2018benchmarking} are generated from 1399 clear images, and we evaluate the performance on 500 indoor images of RESIDE SOTS indoor \cite{li2018benchmarking}. The RESIDE OTS-$\beta$ \cite{li2018benchmarking} consists of 69,510 hazy images generated by 1986 clear images. We evaluate the performance on 500 outdoor images of RESIDE SOTS outdoor \cite{li2018benchmarking}.

The DCPDN-data provided by Zhang and Patel \cite{zhang2018densely}, which consists of total of 4000 training images as TrainA-DCPDN and 400 testing images as TestA-DCPDN \cite{zhang2018densely}. 

The real-world NH-HAZE dataset \cite{ancuti2020nh} has a total of 55 pairs of non-homogeneous hazy and clear images of the real outdoor scenes. We train and evaluate our method according to the NH-HAZE 2020 dehazing challenge \cite{ancuti2020ntire} requirements.

The real-world O-HAZE dataset \cite{ancuti2018haze} consists of a total of 45 pairs of hazy and clear images of the real outdoor scenes. With the O-HAZE 2018 dehazing challenge \cite{ancuti2018ntire} requirements, we follow \cite{9333643} to train and evaluate our method.

The DHQ dataset \cite{min2019quality} contains 250 high-quality real-world hazy images. Moreover, the dataset provides a no-reference quality evaluation method (NR-IQA) \cite{min2019quality} that could be used to evaluate the performance of various dehazing methods in real-world scenes. 

The Foggy Cityscapes-DBF \cite{sakaridis2018model} dataset has the corresponding clear images with bounding box and segmentation masks. We strictly follow \cite{sakaridis2018semantic} to synthesize more kinds of haze density images. Specifically, we set nine kinds of $\beta$, where the value of $\beta$ ranges from 0.004 to 0.02, and the interval is 0.002. Notably, it contains $2975 \times 9$ image pairs for training. Meanwhile, we evaluate on $500\times 9$ images.

\begin{table*}[!h]
	\vspace{-0.2cm}
	\centering
	\caption{full-reference and no-reference quality quality assessment scores on real-world and synthetic dehazing datasets.}
	\resizebox{0.9\linewidth}{!}{
		\begin{tabular}{|c|c|c|c|c|c|c|c|c|c|c|c|c|c|c|c|c|c|}
			\hline
			\multicolumn{2}{|c|}{Method}
			 & DCP \cite{he2010single} & CAP \cite{zhu2015fast} & Meng \textit{et al}. \cite{meng2013efficient} & AMEF \cite{galdran2018image} & DehazeNet \cite{Cai2016DehazeNet} & AOD-Net \cite{li2017aod} & GFN \cite{ren2018gated} & DCPDN \cite{zhang2018densely} & GCANet \cite{chen2019gated} & DuRN \cite{liu2019dual} & GridNet \cite{liu2019griddehazenet} & EDN \cite{yu2020ensemble} & FD-GAN \cite{dong2020fd} & MSBDN \cite{dong2020multi} &MCN \cite{wei2020single} & NHRN\\
			\hline
			\hline
			\multirow{4}*{\shortstack{FQ IQA}}&
			SOTS indoor   & 49.95 & 74.02 & 76.87 & 84.64 & 76.87 & 71.50 & 84.30 & 73.03 & 94.11 & {95.63} & 95.02                   &96.10& 89.07 & 96.66 &\textcolor{blue}{97.22} & \textcolor{red}{97.58}\\
			&SOTS outdoor & 80.51 & 86.47 & 64.96 & 75.87 & 87.18 & 85.39 & 79.95 & 87.39 & 92.95 & 95.02  & {95.09}                 &94.85 & 83.55 & 91.49 &\textcolor{blue}{95.44} & \textcolor{red}{95.87} \\
			&TestA-DCPDN  & 74.42 & 87.93 & 72.99 & 74.63 & 77.34 & 48.02 & 78.14 & 89.90 & 91.94 & {95.63}  &{95.02}               &96.24  & 83.42 & 93.53 &\textcolor{blue}{96.34} &\textcolor{red}{96.45} \\
			& NH-Haze    & 56.92 & 32.79 & 50.32 & 42.04 & 29.02 & 24.72 & 27.65 & 34.68 & 46.24 & 74.86   & \textcolor{blue}{76.59} &72.76& 59.52 & 75.33 & {74.22} & \textcolor{red}{78.68} \\
			&O-HAZE     & 54.10 & 53.52 & 53.32 & 69.67 & 53.84 & 59.35 & 63.28 & 51.92 & 61.99 & \textcolor{blue}{81.97}  & {80.69} & 81.66 & 65.47 & \textcolor{red}{82.94} & {78.14} & {78.34} \\
			\hline
			\hline
			\multirow{1}*{\shortstack{NR IQA}} & DHQ & 48.04 & 45.77 & 41.09 & 48.78 & 48.27 & 48.48 & {49.30} &48.02 & 47.35 &48.29 & 49.09 &49.29& 43.15 & 48.73& \textcolor{blue}{49.36} & \textcolor{red}{49.46} \\
			\hline
	\end{tabular}}
	\vspace{0.2cm}
	\label{result6}
\end{table*}

\begin{figure*}[ht]
    \centering
    \includegraphics[width=1\textwidth]{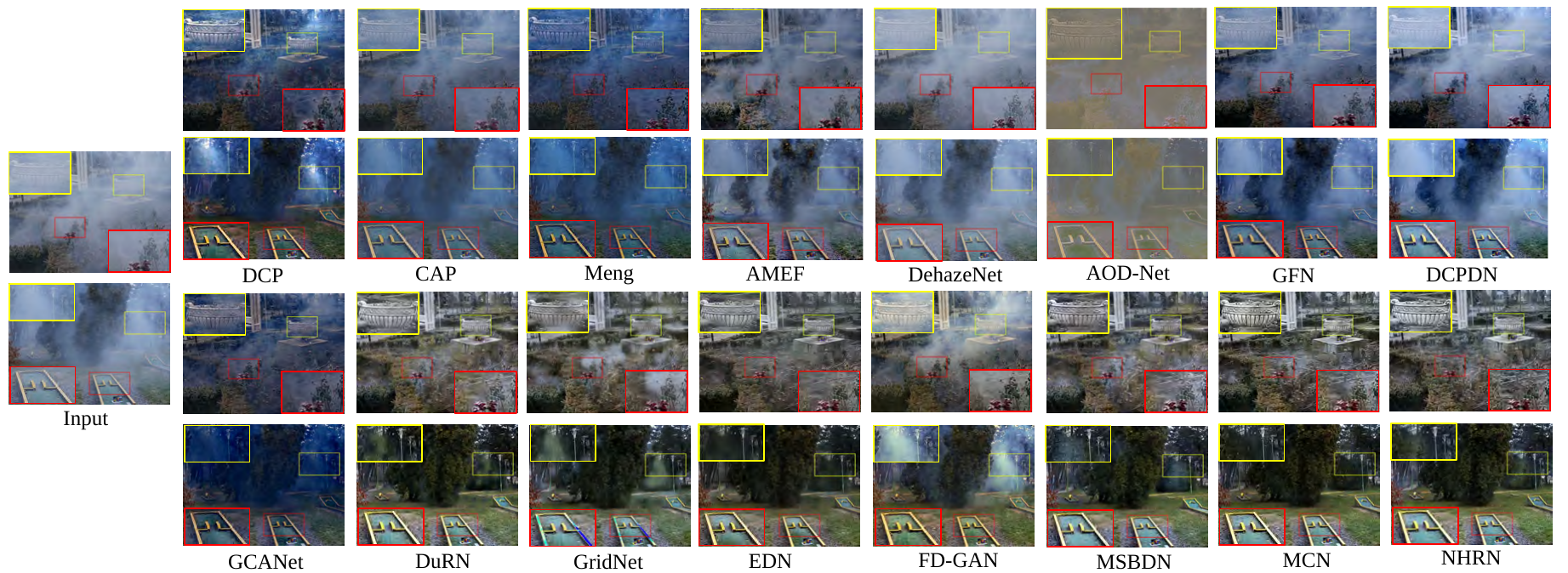}\caption{Qualitative comparison of the state-of-the-art dehazing methods on real- world NH-HAZE dataset.}
    \label{fig:20}
\end{figure*}

\begin{figure*}[ht]
    \centering
    \includegraphics[width=1\textwidth]{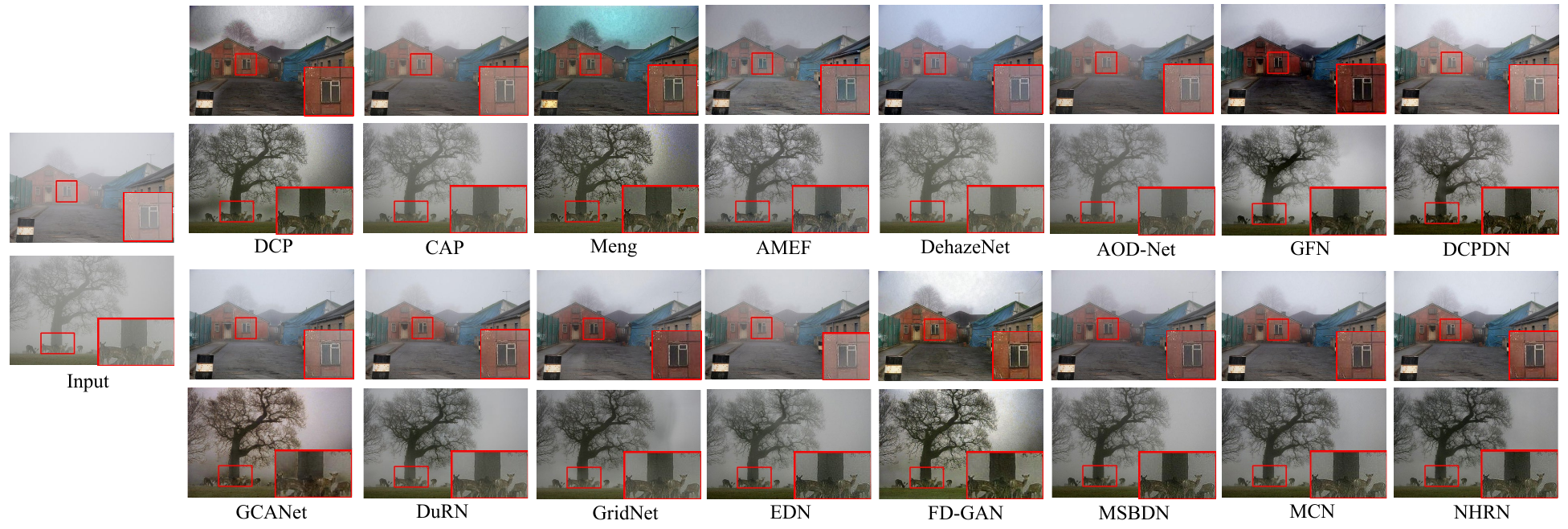}\caption{Qualitative comparison of the state-of-the-art dehazing methods on real- world DHQ dataset.}
    \label{fig:22}
\end{figure*}

\subsection{Implementation Detail}
The network is trained in an end-to-end manner. Specifically, we train the network with the Adam optimizer, where $\beta_{1}$ and $\beta_{2}$ take 0.9 and 0.999, respectively. The initial learning rate is set to 0.0001. We employ a mini-batch size of 16 to train our network. The hazy images are randomly cropped to $224 \times 224$ and used as the input to our network during the training phase. Moreover, we train the network for 120 epochs in the ITS \cite{li2018benchmarking}, TrainA-DCPDN \cite{zhang2018densely}, O-HAZE \cite{ancuti2018haze}, NH-HAZE \cite{ancuti2020nh} decay the learning rate by 0.5 times after 40 epochs. As for OTS-$\beta$, considering a large number of data sets, we train 20 epochs and multiply the learning rate by 0.5 every 10 epochs. Moreover, for the Foggy Cityscapes-DBF \cite{sakaridis2018semantic}, we train 20 epochs with the learning rate of 0.0001. The training is carried out on a PC with NVIDIA GeForce GTX TITAN XP.

\subsection{Quantitative Analysis with Objective Metrics}
Our comparison methods include the traditional dehazing algorithms, e.g., DCP \cite{he2010single}, CAP \cite{zhu2015fast}, Meng \textit{et al}. \cite{meng2013efficient} and AMEF \cite{galdran2018image}, as well as the CNN-based methods, e.g., DehazeNet \cite{Cai2016DehazeNet}, AOD-Net \cite{li2017aod}, GFN \cite{ren2018gated}, DCPDN \cite{zhang2018densely}, GCANet \cite{chen2019gated}, DuRN \cite{liu2019dual}, GridNet \cite{liu2019griddehazenet}, EDN\cite{yu2020ensemble}, FD-GAN \cite{dong2020fd}, MSBDN \cite{dong2020multi} and MCN \cite{wei2020single}. These objective quality metrics, PSNR, {SSIM} (i.e., the higher, the better) and perceptual quality LPIPS \cite{zhang2018unreasonable} (i.e., the lower, the better), are used to measure the dehazing performance. TABLE \uppercase\expandafter{\romannumeral1} reports the quantitative results on synthetic datasets SOTS indoor, SOTS outdoor and TestA-DCPDN, real-world dataset NH-HAZE as well as real-world dataset O-HAZE. It is seen that our proposed method outperforms most the other dehazing methods in terms of PSNR, SSIM, and LPIPS metrics. Only on the DCPDN-data and O-HAZE
datasets, our NHRN is slightly inferior to EDN and GCANet in terms of the LPIPS metric, whose gaps are just 0.003 and 0.056, respectively. Furthermore, our improved method NHRN performs better than our previous method \cite{wei2020single} in terms of PSNR, SSIM and LPIPS metrics on five datasets, especially for the NH-HAZE dataset. The results of competing methods mainly refer to \cite{li2020deep, 9333643, deng2019deep, liu2019dual, liu2019griddehazenet, yu2020ensemble}. The other unavailable data in the aforementioned literature are obtained by retraining the models released by the authors.

\begin{figure*}[ht]
    \centering
    \includegraphics[width=0.9\textwidth]{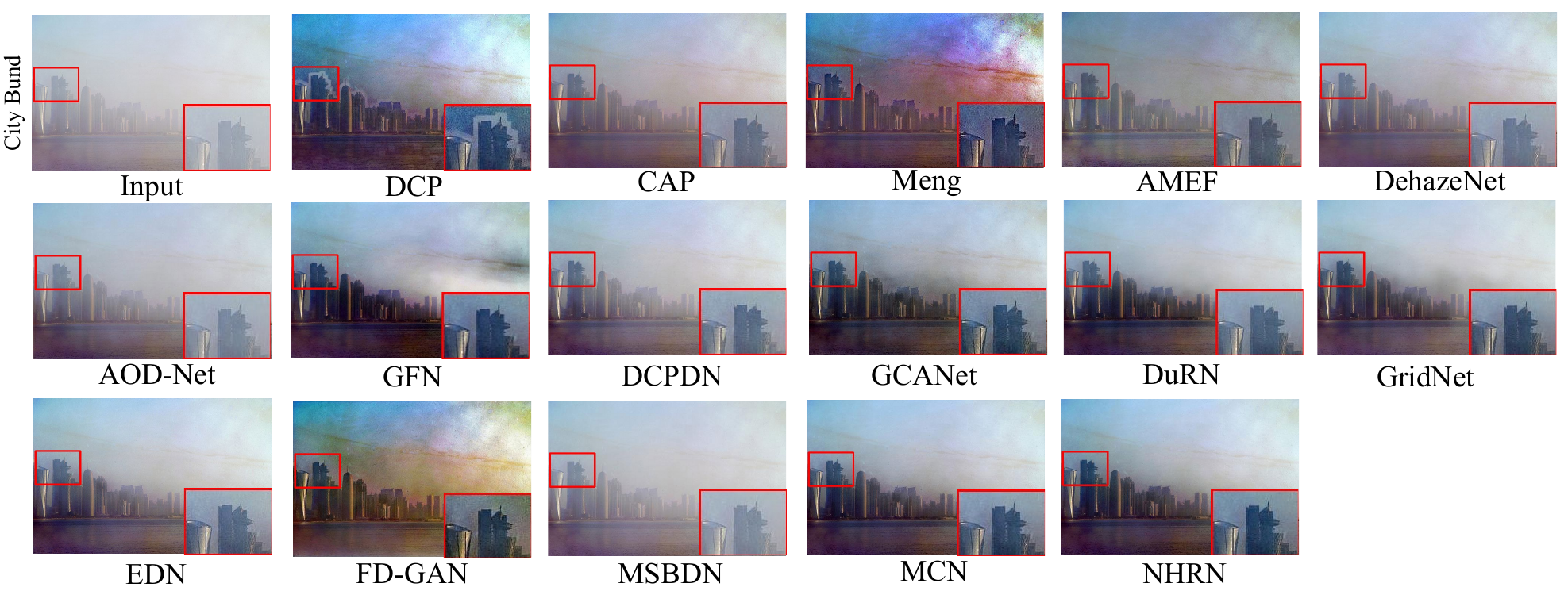}\caption{Qualitative comparison of the state-of-the-art dehazing methods on the  city bund.}
    \label{fig:23}
\end{figure*}

\subsection{Quantitative Analysis with IQA Metrics}
In this part, we introduce the image quality assessment (IQA) metrics to compare different dehazing methods as a complement to three metrics, PSNR, SSIM, and LPIPS. Specifically, we adopt a full-reference quality assessment (FR-IQA) \cite{min2019quality} to evaluate SOTS indoor, SOTS outdoor, TestA-DCPDN, NH-HAZE, and O-HAZE. In the FR-IQA evaluation algorithm, we follow the method of Min \textit{et al.}, which extracts three groups of dehazing image features to predict their scores via support vector regression or random forest.

As a further experiment supplement, we apply NR-IQA \cite{min2018objective} as a no-reference quality assessment to evaluate the dehazing effect in real-world hazy images. In the NR-IQA evaluation algorithm \cite{min2018objective},
Min \textit{et al.} integrates various haze-relevant features, which conducts an effective subjective quality evaluation.

As seen in TABLE \uppercase\expandafter{\romannumeral2}, for FR-IQA metric, our method shows sufficient advantages on different datasets. In particular, our method achieves the best performance, whose score is 97.58 on SOTS indoor, 95.87 on SOTS outdoor, 96.45 on TestA-DCPDN, 78.68 on NH-HAZE, and 78.34 on O-HAZE dataset, respectively. Moreover, we present the results of NR-IQA for the DHQ dataset. Our method achieves the best score of 49.46, where the gaps with our previous method \cite{wei2020single} is 0.1.

\subsection{Qualitative Evaluation}
We further show the dehazing results of the state-of-the-art dehazing methods on real-world NH-HAZE and DHQ datasets in Fig. 8 and Fig. 9 for qualitative comparisons. It is observed that our method achieves desirable dehazing results on these datasets, which show the robustness of our method. From visual results, we can observe that DCP \cite{he2010single}, CAP \cite{zhu2015fast} and Meng \textit{et al.} \cite{meng2013efficient} easily suffer from color distortion, which makes the brightness of several areas relatively dark. AMEF \cite{galdran2018image}, DehazeNet \cite{Cai2016DehazeNet}, AOD-Net \cite{li2017aod} and DCPDN \cite{zhang2018densely} still remain haze in heavily hazy scene. The processing power of GFN \cite{ren2018gated}, GCANet \cite{chen2019gated} and FD-GAN \cite{dong2020fd} on the high-frequency detail information performance is unnatural. Although DuRN \cite{liu2019dual}, GridNet \cite{liu2019griddehazenet}, EDN \cite{yu2020ensemble} and MSBDN \cite{dong2020multi} look good in some cases, there are flaws in detail restoration. Our method achieves the best visual effect, which could both preserve the image detail and color as well as remove the haze as much as possible from the input. This is because our method could provide suitable scene prior. Meanwhile, our method conducts non-local filtering in the spatial regions and channel dimensions to model long-range dependency and propagate information between graph nodes for better dehazing optimization.

\begin{table}[!t]
	\vspace{-0.2cm}
	\centering
	\caption{Qualitative comparisons on the images (City Bund) in fig. 10 based on the indicators $\textit{e}$, $\Sigma$,  $\overline{r}$, and FADE of Hautière \textit{et al.} \cite{hautiere2008blind} and choi \textit{et al.} \cite{choi2015referenceless}}
	\resizebox{0.5\linewidth}{!}{
		\begin{tabular}{|c|c|c|c|c|c|c|c|c|c|c|}
			\hline
			& \multicolumn{4}{|c|}{City Bund}\\
			\hline
			Method & $\textit{e}$ $\uparrow$ & $\Sigma$ $\downarrow$ &  $\overline{r}$ $\uparrow$  & FADE $\downarrow$ \\
			\hline
			\hline
			DCP   \cite{he2010single}         & \textcolor{blue}{9.899} & 0 & \textcolor{blue}{4.110} &\textcolor{blue}{0.428}  \\
			CAP \cite{zhu2015fast}          & 2.353 & 0 & 1.363 & 1.397 \\
			Meng \cite{meng2013efficient}   & \textcolor{red}{11.556} & 0 & \textcolor{red}{4.626} & \textcolor{red}{0.404} \\
			AMEF \cite{galdran2018image}   & 3.291 & 0 & 2.255 & 1.424 \\
			\hline
			\hline
			DehazeNet \cite{Cai2016DehazeNet}     & 1.852 & 0 & 1.525 & 2.059  \\
			AOD-Net \cite{li2017aod}       & 1.435 & 0 & 1.497 & 3.189 \\
			GFN \cite{ren2018gated}       & 4.276 & 0 & 2.348 & 0.873 \\
			DCPDN \cite{zhang2018densely}        & 2.496 & 0 & 1.954 & 2.123 \\
			GCANet \cite{chen2019gated}         & 3.685 & 0 & 2.195 & 1.353  \\
			DuRN  \cite{liu2019dual}         & 4.008 & 0 & 2.093 & {1.765}  \\
			GridNet \cite{liu2019griddehazenet}  & {3.347} & {0} & {1.939} & 2.144 \\
			EDN \cite{yu2020ensemble}  & {2.940} & {0} & {1.806} & 2.231 \\
			FD-GAN \cite{dong2020fd}  & {7.441} & {0} & {3.420} & 0.537  \\
			MSBDN \cite{dong2020multi}  & {0.855} & {0} & {1.407} & 3.749  \\
			\hline
			\hline
			\textbf{MCN}\cite{wei2020single}  & {4.258} & {0} & {2.327} & 1.361 \\
			\textbf{NHRN}  & {4.407} & {0} & {2.333} & 1.261 \\
			\hline
	    \end{tabular}}
	\vspace{0.2cm}
	\label{result3}
\end{table}
\subsection{Quantitative Analysis with Blind Measures}
Fig. 10 shows qualitative comparisons against the recent dehazing methods. We refer to these blind measures in this review \cite{singh2019comprehensive}. Specifically, we employ indicators $\textit{e}$, $\Sigma$, $\overline{r}$ and FADE as main blind measures, which focus on evaluating the edge visibility, image contrast, the percentage of extreme dark/bright pixels, and the haze density after dehazing. The higher $\textit{e}$, $\overline{r}$ and lower $\Sigma$, FADE are considered as better dehazing results. As shown in TABLE \uppercase\expandafter{\romannumeral3}, Meng \textit{et al}. outperforms the other methods in terms of these four blind measures. It should be noted that the aforementioned blind measures only focus on specific attributes of the enhanced image, which are hard to simulate complex perceptual characteristics of the human vision system. As shown in Fig. 10, Meng \textit{et al}. over-enhances the color and contrast of the haze image. The proposed NHRN performs better in balancing the haze removal and detail restoration.

\begin{table}[t]
	\vspace{-0.2cm}
	\caption{Detection results on the Foggy Cityscapes-DBF with nine haze concentrations.}
	\centering
	\resizebox{1\linewidth}{!}{
		\begin{tabular}{|c|c|c|c|c|c|c|c|c|c|}
			\hline
			& \multicolumn{9}{|c|}{mAP and AP$_{50}$ on Foggy Cityscapes-DBF test set (mAP/AP$_{50}$)} \\
			\hline
			Method & $\beta = 0.004$ & $\beta = 0.006$ & $\beta = 0.008$ & $\beta = 0.01$ & $\beta = 0.012$ &  $\beta = 0.014$ & $\beta = 0.016$ &  $\beta = 0.018$ & $\beta = 0.02$ \\
			\hline
			\hline
			baseline  & 33.6/54.6 & 31.3/50.6 & 28.8/46.1 & 26.2/40.8 & 24.4/38.2 & 23.3/36.2 & 22.0/34.0 & 20.3/31.8 & 19.1/29.9 \\
			\hline
			DCP \cite{he2010single}  & 32.3/53.9 & 31.1/52.9 & 29.7/50.7 & 28.0/46.9 & 26.9/44.8 & 25.9/43.2 & 24.4/40.7 & 23.2/38.8 & 22.3/37.1 \\
			CAP \cite{zhu2015fast}  & 35.6/57.9 & 34.2/56.4 & 32.6/53.8 & 31.2/51.1 & 29.8/49.2 & 28.2/45.9 & 26.4/ 41.7 & 25.5/40.3 & 24.2/37.9 \\
			Meng \textit{et al.} \cite{meng2013efficient}  & 36.1/58.7 & 35.4/57.2 & 34.4/55.2 & 33.4/53.7 & 32.1/52.3 & 31.4/51.1 & 30.2/49.5 & 29.3/48.1 & 28.4/46.2 \\
			AMEF \cite{galdran2018image} & 33.9/56.0 & 32.6/54.1 & 31.2/51.3 & 29.8/48.9 & 28.7/47.0 & 27.2/44.6 & 26.3/42.9 & 25.1 /40.2&24.0/38.1 \\
			\hline
			\hline
			DehazeNet \cite{Cai2016DehazeNet} & 29.4/49.4 & 30.1/49.5 & 29.8/49.7 & 28.9/48.2 & 27.8/46.7& 26.9/44.5 & 25.4/41.2 & 24.6/39.0& 23.8/37.4 \\
			AOD-Net \cite{li2017aod}  & 32.8/54.0 & 30.5/50.2  &29.4/48.7 & 28.0/47.0 & 27.4/46.1 & 26.1/44.3 & 25.3/42.4 & 24.6/40.9 & 23.9/39.3 \\
			GFN \cite{ren2018gated} & 28.0/46.9 &26.5/44.7 & 24.7/41.2 & 23.3/39.2 & 22.1/35.8 & 20.5/33.5 & 19.5/31.6 & 18.2/30.8 & 17.1/29.0 \\
			DCPDN \cite{zhang2018densely} & 34.1/55.0 &31.8/51.5 & 29.5/47.5 &27.0/43.4 & 25.1/40.2 & 23.6/37.3 & 22.2/34.8 & 21.0/32.7 & 19.8/30.7 \\
			GCANet \cite{chen2019gated} & 34.0/57.2 & 33.1/54.8 & 32.2/53.3 & 31.6/52.7 & 31.1/51.0 &30.4/50.4 &29.5/48.8&28.7/47.8 & 28.1/46.9 \\
			DuRN \cite{liu2019dual} & 37.5/60.9 & 36.7/59.5 & 35.5/58.0 & 34.6/55.9 & 33.3/54.2 & 32.1/52.8 & 31.5/51.0 & 30.0/48.6 & 29.1/47.0\\
			GridNet \cite{liu2019griddehazenet}  & 37.4/60.8 &36.6/60.0 & 35.5/58.6 & 34.6/57.3 &33.5/55.2 & 32.5/53.6 &31.2/52.1 & 30.2/50.9 &29.6/49.1\\
			EDN \cite{yu2020ensemble}    & 36.0/58.9 &35.4/58.5 & 35.3/57.8 & 34.3/56.6 & 33.8/55.5 & 33.2/54.0 &32.5/52.5 & 31.8/51.4 & 31.5/51.9\\
			FD-GAN \cite{dong2020fd} & 28.3/48.7 & 27.2/46.6 & 25.9/45.5 & 24.5/42.3 & 23.6/40.1 & 22.8/38.6 & 21.9/37.0 & 21.2/35.4 & 19.9/33.3\\
			MSBDN \cite{dong2020multi} & 34.5/56.6 & 33.4/55.7 & 32.0/53.0 & 30.4/50.2 & 28.3/47.3 & 26.4/44.4 & 25.3/42.2& 24.3/40.5 & 22.7/37.6\\
			\hline
			\hline
		    MCN \cite{wei2020single}  & 38.2/62.1 &37.2/61.1 & 36.0/59.4 & 35.5/58.5 &34.6/56.2 & 33.8/56.0 &32.9/54.1 & 32.2/52.6 &31.1/51.6\\
			\textbf{NHRN} & \textbf{38.5/62.6} & \textbf{37.7/61.9} & \textbf{37.0/60.9} & \textbf{36.4/59.3} & \textbf{35.7/58.9} & \textbf{35.4/57.1}  & \textbf{34.4/56.2} & \textbf{33.5/54.0} &
			\textbf{32.6/53.1} \\
			\hline
			Ideal case  & \multicolumn{9}{|c|}{39.8/64.5} \\
			\hline
	\end{tabular}}
	\vspace{0.2cm}
	\label{result4}
\end{table}

\begin{table}[t]
	\vspace{-0.2cm}
	\caption{Segmentation results on the Foggy Cityscapes-DBF with nine haze concentrations.}
	\centering
	\resizebox{1\linewidth}{!}{
		\begin{tabular}{|c|c|c|c|c|c|c|c|c|c|}
			\hline
			& \multicolumn{9}{|c|}{mIoU and mAcc on Foggy Cityscapes-DBF test set (mIoU/mAcc)} \\
			\hline
			Method & $\beta = 0.004$ & $\beta = 0.006$ & $\beta = 0.008$ & $\beta = 0.01$ & $\beta = 0.012$ &  $\beta = 0.014$ & $\beta = 0.016$ &  $\beta = 0.018$ & $\beta = 0.02$ \\
			\hline
			\hline
			baseline  & 72.72/79.25 & 69.74/76.13 & 66.62/72.96 & 63.71/70.00 & 61.01/67.27 & 58.30/64.54 & 55.83/62.08 & 53.45/59.74 & 51.06/57.37 \\
			\hline
			DCP \cite{he2010single}  & 52.02/63.86 & 50.96/62.71 & 49.58/61.36 & 48.43/60.16 & 47.31/59.07 & 46.03/57.88 & 44.82/56.86 & 43.49/55.80 & 42.14/54.70 \\
			CAP \cite{zhu2015fast}  & 68.67/79.09 & 68.12/78.42 & 67.37/77.56 & 66.45/76.52 & 65.64/75.52 & 64.43/74.35 & 63.57/73.26 & 61.88/71.81 & 60.76/70.51 \\
			Meng \textit{et al}. \cite{meng2013efficient} & 66.05/77.81 & 65.54/77.03 & 65.11/76.29 & 64.56/75.47 & 63.94/74.58 & 63.26/73.69 & 62.60/72.84 & 61.92/72.05 & 61.32/71.32 \\
			AMEF \cite{galdran2018image} & 69.18/76.22 & 67.78/74.62 & 66.49/73.20 & 65.29/71.86 & 63.94/70.43 & 62.58/69.05 & 61.22/67.73 & 60.01/66.56&58.60/65.14 \\
			\hline
			\hline
			DehazeNet \cite{Cai2016DehazeNet}& 37.87/50.47 & 40.36/52.77 & 42.27/54.45  & 44.08/56.09 & 45.13/56.92 & 46.26/57.87 & 47.07/58.59 & 47.51/58.87& 47.70/58.88 \\
			AOD-Net \cite{li2017aod} & 66.61/76.35 & 64.03/73.59 &61.78/71.04  & 59.61/68.68 & 57.56/66.46 & 55.58/64.31 & 53.58/62.09 & 51.65/59.97& 49.66/57.86 \\
			GFN \cite{ren2018gated}& 55.78/68.03 & 54.78/66.68 & 53.04/64.65 & 51.27/62.74 & 49.56/60.74 & 48.01/58.90 & 46.36/57.05 & 44.67/55.18 & 43.23/53.52 \\
			DCPDN \cite{zhang2018densely} & 64.29/75.00 & 61.21/71.95 & 57.83/68.59 & 55.05/65.57 & 52.48/62.82 & 50.00/60.26 & 47.58/57.83 & 45.41/55.71 & 43.31/53.66 \\
			GCANet \cite{chen2019gated}  & 64.51/73.45 & 63.31/72.12 & 61.66/70.55 & 60.15/69.10 & 58.69/67.69 &57.27/66.35 & 55.94/65.17 & 54.68/63.94 & 53/36/62.86 \\
			DuRN \cite{liu2019dual} & 75.31/82.22 & 74.29/81.15 & 73.22/80.07 & 72.20/78.94 & 71.09/77.82 & 70.03/76.66 & 68.86/75.38 & 67.70/74.13 & 66.40/72.81\\
			GridNet \cite{liu2019griddehazenet} & 74.65/81.85 & 73.82/81.00 & 72.99/80.04 & 72.07/79.01 & 71.03/77.90 & 69.79/76.74 &68.59/75.47 & 67.23/74.17&65.86/72.87\\
			EDN \cite{yu2020ensemble}    & 65.67/74.51 &67.23/75.31 & 67.83/75.36 & 67.88/75.16 & 67.96/74.87 & 67.69/74.29 &67.52/73.95 & 67.21/73.48& 66.60/72.73\\
			FD-GAN \cite{dong2020fd} & 60.15/71.07 & 59.17./70.11 & 58.17/69.23 & 57.11/68.29 & 56.04/67.27 & 54.77/66.09 & 53.85/65.15 & 52.72/64.07 & 51.86/63.18\\
			MSBDN \cite{dong2020multi} & 67.00/74.33 & 65.10/72.53 & 62.92/70.55 & 60.58/68.45 & 58.32/66.35 & 55.79/63.93 & 53.26/61.41 & 50.84/59.00& 48.50/56.71\\
			\hline
			\hline
			MCN \cite{wei2020single} & 75.86/82.55 & 75.26/81.83 & 74.49/80.95 & 73.51/79.88 & 72.61/78.99 &71.17/77.58 & 69.84/76.23&68.65/75.05 &67.26/73.71\\
			\textbf{NHRN} & \textbf{76.50/83.41} & \textbf{76.13/82.94} & \textbf{75.70/82.45} & \textbf{75.17/81.84} & \textbf{74.46/81.11} & \textbf{73.73/80.29}  & \textbf{73.01/79.51} & \textbf{72.23/78.69} &
			\textbf{71.42/77.83} \\
			\hline
			Ideal case  & \multicolumn{9}{|c|}{77.30/84.30} \\
			\hline
	\end{tabular}}
	\vspace{0.2cm}
	\label{result}
\end{table}

\begin{figure*}[ht]
    \centering
    \includegraphics[width=0.9\textwidth]{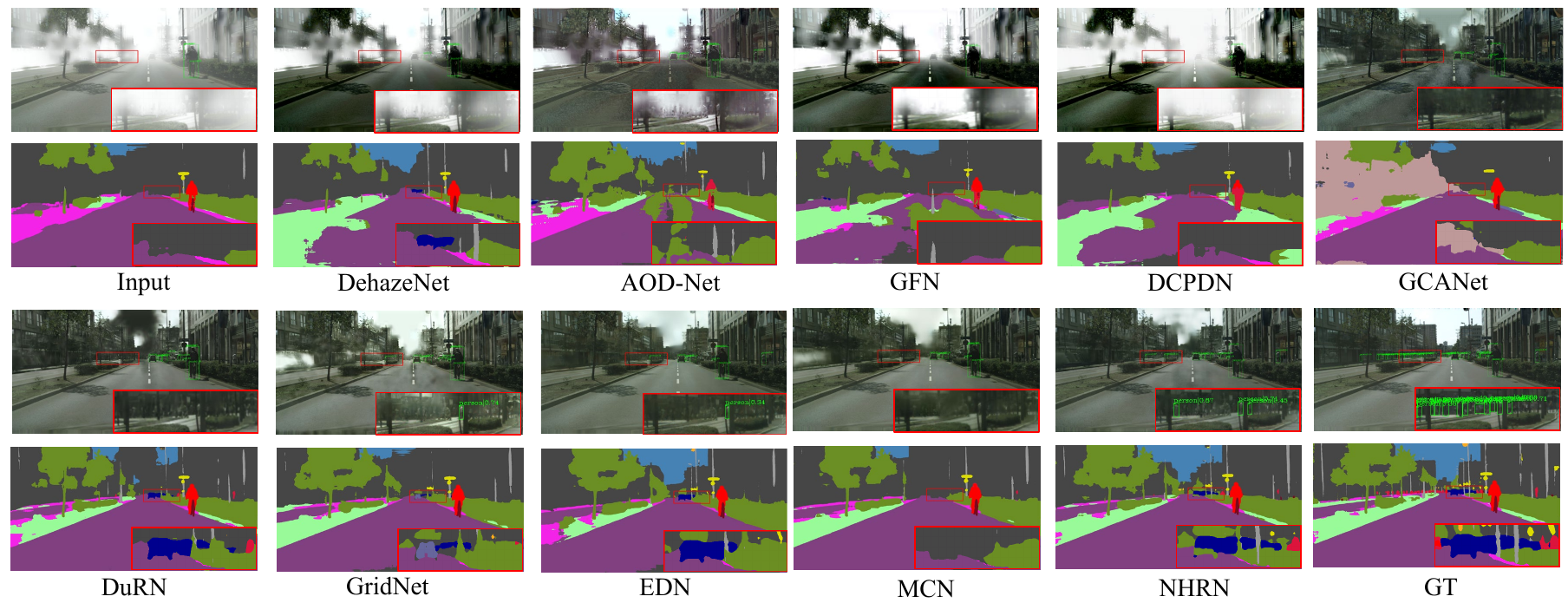}
    \caption{Visualization of detection and segmentation under Foggy Cityscapes-DBF dataset.}
    \label{fig:9}
\end{figure*}

\begin{figure*}[t]
    \centering
    \includegraphics[width=0.65\textwidth]{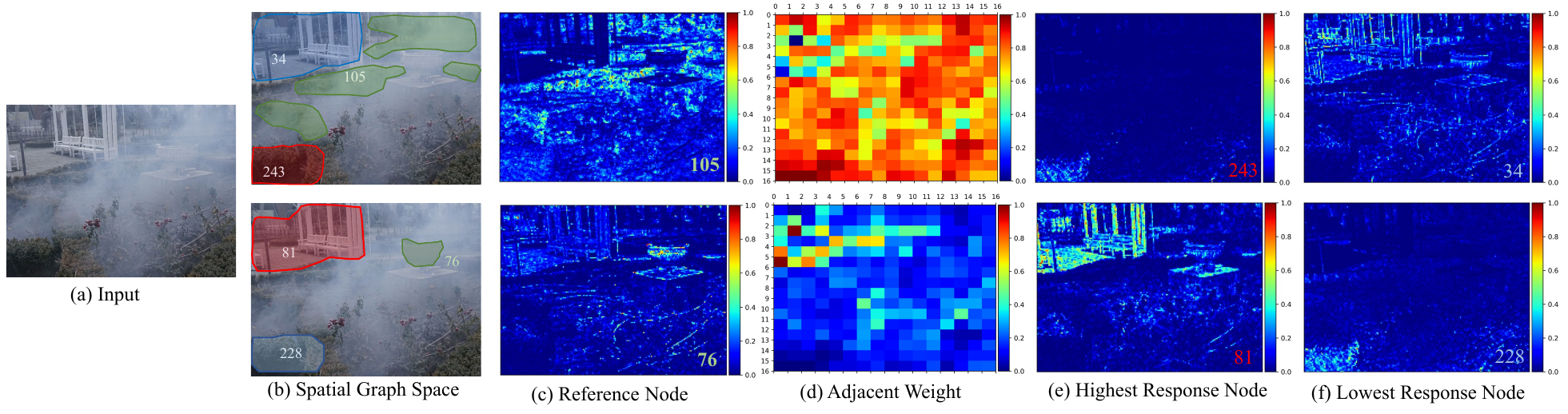}\caption{Visualization of the learning projection weights and adjacency weights on the spatial graph nodes (16*16 spatial nodes). We show the spatial graph Node-105, 76 on the first and second rows, respectively. Specifically, (a) is a non-homogeneous hazy image. Column (b) shows the illustration of graph space. Column (c) shows the learning projection weights on the Node-105 and 76. Column (d) shows the adjacent weights on the Node-105 and 76. Besides, column (e) and (f) show the projection weights of the response nodes that have the highest and lowest adjacency weights for Nodes-105 and 76, respectively. In particular, Node-105 lends to aggregate grass under the dense haze. the Node-243 represents the clear regions of grass, and propagates the most interactive information to the Node-105. Meanwhile, Node-76 mainly focuses on the parterre under dense haze, with its most interactive information coming from Node-81. In particular, Node-81 mainly consists of the pixel-wise features of the pavilion in the clearer region.}
    \label{fig:10}
\end{figure*}

\begin{figure*}[t]
    \centering
    \includegraphics[width=0.65\textwidth]{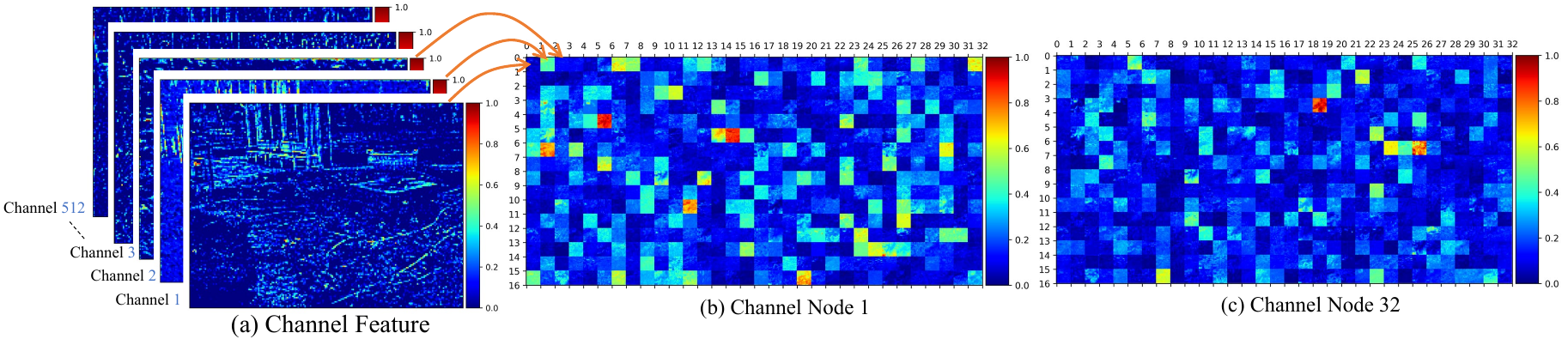}\caption{Visualization of the learning adjacency weights on the channel {graph} nodes (512 channel nodes). Specifically, (a) is the channel feature. Furthermore, we splice the 512 channel images according to 32 $\times$ 16, and add corresponding adjacent weights. In particular, (b) and (c) are the adjacency matrices of reference channel Node-1 and 32, respectively.}
    \label{fig:11}
\end{figure*}

\subsection{Effect of Haze Scene Detection and Segmentation}
In addition to visual effectiveness, we further compare the performance of these dehazing algorithms in detection and segmentation tasks using the Foggy Cityscapes-DBF dataset. For object detection task, we adopt Faster-RCNN \cite{ren2016faster} as our detector, which is pretrained on the Cityspaces dataset \cite{cordts2016cityscapes}. We employ mAP and AP$_{50}$ to measure the detection effect. As shown in TABLE \uppercase\expandafter{\romannumeral4}, our algorithm achieves the highest detection precision in the dehazing results of nine haze concentrations. More intuitively, Fig. 11 shows that our method has lesser missed detection and is close to the target detection result under the haze-free scene compared to other CNN-based methods. Meanwhile, we conduct similar experiments on the semantics segmentation task, whose metric consists of mIoU and mAcc. We exploit PSPNet \cite{zhao2017pyramid} as the semantics segmentation algorithm pretrained on the Cityspaces dataset. The comparison results are shown in TABLE \uppercase\expandafter{\romannumeral5}. As revealed in Fig. 11, it can be observed that our method leads to more consistent semantic information with ground-truth on the qualitative segmentation results. Note that conducting object detection and segmentation directly on hazy images and haze-free images are adopted as the baseline and ideal case, respectively.

\begin{table*}[htbp]
\vspace{-0.2cm}
\caption{Performance comparison on SOTS for the number of nodes  of SGR.}

\begin{center}
\resizebox{0.8\linewidth}{!}{
\begin{tabular}{|c||c||c||c||c||c||c||c||c||c||c||c||c||c|}
\hline
nodes  &1*1     &2*2     & 4*4    & 6*6    & 8*8    & 10*10  & 12*12  &14*14   & 16*16           &18*18  &20*20     &22*22 & 24*24  \\
\hline
PSNR               & 35.11  & 35.14  & 35.28  & 35.38  & 35.36  & 35.49  & 35.55  &35.47   &\textbf{35.60}   &35.56   & 35.45   &35.26 & 35.51 \\
\hline
SSIM               & 0.9876 & 0.9881 & 0.9878 & 0.9882 & 0.9883 & 0.9882 & 0.9885 &0.9882  &\textbf{0.9886}  &0.9883  &{0.9880} &0.9881& 0.9884 \\
\hline

\end{tabular}}
\vspace{0.2cm}
\label{tab:5}
\end{center}
\end{table*}

\begin{table}[htbp]
    \vspace{-0.1cm}
      \caption{Comparisons on SOTS for different configurations.}
    \centering
    \resizebox{0.85\linewidth}{!}{
    \begin{tabular}{|c||c||c||c||c||c||c||c||c||c|}
    \hline
    AMS &   &$\surd$&$\surd$ &       &      &         & $\surd$ & $\surd$ &$\surd$ \\
    GF &   &       &$\surd$ &       &      &         & $\surd$ & $\surd$ &$\surd$ \\
    SGR &   &       &        &$\surd$&      & $\surd$ & $\surd$  &         &$\surd$\\
    CGR &   &       &        &      &$\surd$& $\surd$ &         & $\surd$ &$\surd$\\
    \hline
    PSNR &  33.14 & 34.70 & 35.06 & 35.33 & 35.09  & 35.40 &  35.48 &  35.26  &\textbf{35.60}\\
    SSIM & 0.9864 & 0.9878 & 0.9885 & 0.9883 & 0.9881 & 0.9882 &  0.9885  &  0.9883 &\textbf{0.9886}\\
    Parameters & 28.87M & 28.87M & 28.97M & 29.66M & 29.53M  & 30.31M &  29.76M &  29.63M  & 30.42M\\
    FLOPs & 61.68G & 62.55G & 88.24G & 64.91G & 63.37G  & 66.60G &  92.02G &  90.47G  & 93.71G\\
    \hline
    \end{tabular}}
    \vspace{0.1cm}
    \label{tab:6}
\end{table}

\begin{table}[h]
\vspace{-0.2cm}
\caption{Performance comparison on SOTS for some pre-trained models.}
\begin{center}
\resizebox{0.4\linewidth}{!}{
\begin{tabular}{|c||c||c|}
\hline
Pre-trained models  &PSNR&SSIM  \\
\hline
VGG-16             &22.08 & 0.8905 \\
\hline
VGG-19             &22.45 &0.9844  \\
\hline
Resnet-18             &33.75  & 0.9868 \\
\hline
Resnet-34           &33.74 & 0.9886 \\
\hline
Resnet-50           &\textbf{35.60} & \textbf{0.9886}  \\
\hline
Resnet-101          &35.49 & 0.9884  \\
\hline
\end{tabular}}
\vspace{0.2cm}
\label{tab:7}
\end{center}
\end{table}

\begin{table}[h]
\vspace{-0.2cm}
\caption{Performance comparison on SOTS for the hyper-parameters $\lambda$.}
\begin{center}
\resizebox{0.3\linewidth}{!}{
\begin{tabular}{|c||c||c|}
\hline
 $\lambda$ & PSNR &SSIM \\
\hline
0             &34.58 & 0.9787 \\
\hline
0.005           &34.97 & 0.9847  \\
\hline
0.01           &35.18 & 0.9861\\
\hline
0.05          &35.31 & 0.9872 \\
\hline
0.1           &\textbf{35.60} & \textbf{0.9886} \\
\hline
0.5           &35.50 & 0.9884 \\
\hline
1           &35.37& 0.9885 \\
\hline
5           &34.90& 0.9884 \\
\hline
10           &34.87& 0.9883 \\
\hline
\end{tabular}}
\vspace{0.2cm}
\label{tab:8}
\end{center}
\end{table}

\subsection{Analysis and Visualization}
We consider that our method mainly relies on the proposed SGR and CGR module, which is able to capture the underlying non-local contextual information. Therefore, we first visualized the SGR module learning the projection weights and corresponding adjacent weights (see Fig. 12). In particular, we totally set $16^{2}$ spatial graph nodes. The projection weights (i.e., $B_{s}$ in $Eq. (14)$) could reflect the irregular region formed by the feature aggregation of the spatial graph nodes. The adjacent weight diffuses information across nodes, where the higher weight represents stronger relations from the current node to the reference node, while the lower weight indicates weak relations to the reference node. We show the spatial graph Node-105, 76 on the first and second rows, respectively. We can see that Node-105 lends to focus on the pixel-wise features of grass under the dense haze, while Node-76 tends to aggregate the pixel-wise features of the parterre under the dense haze. Column (d) shows the adjacent weights on the Node-105 and 76. Besides, column (e) and (f) show the projection weights of the response nodes that have the highest and lowest adjacency weights for Nodes-105 and 76, respectively. We observe that Node-105 has been able to interact with long-range Node-243, which aggregates the clear regions of grass. Meanwhile, Node-76 captures the non-local contextual information from Node-81, which focuses on the pavilion in the clearer region with similar structures.

\begin{figure}[t]
    \centering
    \includegraphics[width=0.3\textwidth]{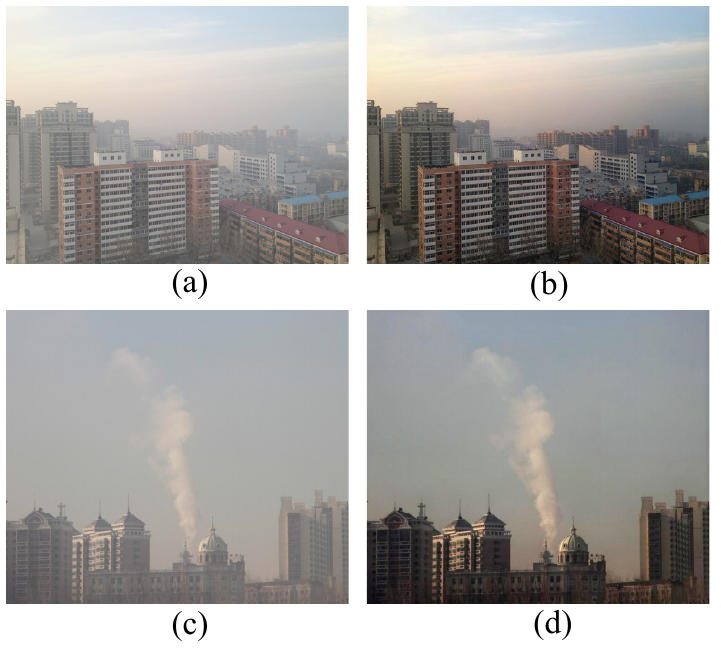}
    \caption{Visualization of our dehazing performance under non uniform atmospheric light (a) and white bright objects (c). (b) and (d) are the dehazing results for (a) and (c), respectively.}
    \label{fig:28}
\end{figure}

For CGR module, we provide the corresponding adjacent weights added on each channel feature map (see Fig. 13). Furthermore, we could observe that channel Node-1 and 32 could capture the interaction with non-local channel nodes. Meanwhile, each reference node could learn an adaptive adjacency matrix according to its dependencies with other nodes. Therefore, these visualizations demonstrate that SGR and CGR could build long-range dependencies and propagate the natural scene prior between the well-preserved nodes and the nodes contaminated by haze.

\subsection{Effect of Complex Scene}
{Fig. 14 shows more visual results under non-uniform atmospheric light and white bright objects like cloud, fog, etc. We can see that our method performs well in coping with the haze images with complex atmospheric light, which removes the non-homogeneous haze and well preserves the texture and color of the cloud. The dehazing performance benefits from the proposed scene prior and bidimensional graph reasoning.}

\subsection{Ablation Study}
\subsubsection{Effect of the components}
We perform the ablation studies to verify the major components of our network. As shown in TABLE \uppercase\expandafter{\romannumeral7}, overall, the PSNR and SSIM gradually rise when feeding AMS, GF, SGR, and CGR. From the charts of PSNR and SSIM indicators, we can see that these two metrics increase when adding the AMS. When adding the SGR and CGR, PSNR and SSIM are significantly improved relative to the basic network.  It confirms the benefit of reasoning across different spatial regions and multiple artificial shots for single image dehazing.

\subsubsection{The number of spatial graph nodes}
We perform the ablation studies to analyze the effect of different numbers of spatial graph nodes on the results. As shown in TABLE \uppercase\expandafter{\romannumeral6}, the PSNR and SSIM rise to the best performance when setting the number of nodes is $16\times16$. Meanwhile, increasing the number of nodes further does not bring performance improvement, which is because more detailed anchors affect the overall feature representation.

\subsubsection{Effect of the pre-trained models}
{We investigate the effect of different pre-trained models as the encoder network for the SOTS indoor dataset. TABLE} \uppercase\expandafter{\romannumeral8} {shows that the Resnet \cite{he2016deep} networks as the pre-trained models perform better than Vgg \cite{simonyan2014very} network. The advantages of Resnet networks mainly benefit from the residual learning framework to get the deeper network. When we increase network depth from 18 layers to 50 layers, PSNR and SSIM both gradually rise. When we continue to increase network depth to 101 layers, the PSNR and SSIM begin to decrease a little. Based on this investigation, we employ Resnet-50 as the pre-trained models in our experiment.}

\subsubsection{Effect of the hyper-parameters}
{We investigate the influences of different settings of hyper-parameters $\lambda$ for $Eq. (17)$. The evaluation results are shown in TABLE} \uppercase\expandafter{\romannumeral9}. {When $\lambda$ is increased from 0 to 0.1, PSNR and SSIM are significantly improved relatively. However, when we continue to increase $\lambda$, the PSNR and SSIM decrease a little. The best dehazing performance is obtained by setting $\lambda$ to 0.1.}

\subsubsection{Complexity of the proposed model}
 {TABLE} {\uppercase\expandafter{\romannumeral7}} {shows the complexity of the proposed model. It can be clearly seen that the total parameter and FLOPs (Floating-point Operations on the 512$\times$512 image size) of the proposed model is 30.42M and 93.71G. Compared with the baseline, AMS, GF, SGR, and CGR only bring a small increase in the number of parameters. In terms of FLOPs, GF produces a lot of complexity, mainly because GF calculates weights map on high resolution features. Meanwhile, SGR and CGR benefit from the compact representation of the graph, which only brings a small computational cost.}

\subsubsection{The number of AMS}
We perform the experiment by considering different amounts of artificial shots on the basis of our baseline network with GF module, where GF module is a compensation for AMS. As shown in Fig. 15, overall, the PSNR and SSIM gradually rise when the number of artificial shots increases from 0 to 4. From the charts of PSNR and SSIM indicators, it can be seen that there is a slight increase when adding one artificial shot. When the number is increased form 2 to 4, PSNR and SSIM is significantly improved relative to no artificial shot. However, when we continue to increase the number of shots, the PSNR and SSIM decrease a little. That is because the continuing feed in high-frequency compensation brings unnecessary noise interference. It confirms the benefit of appropriately enriching high-frequency input for single image dehazing.

\begin{figure}[t]
    \vspace{-0.2cm}
    \centering
    \includegraphics[width=0.35\textwidth]{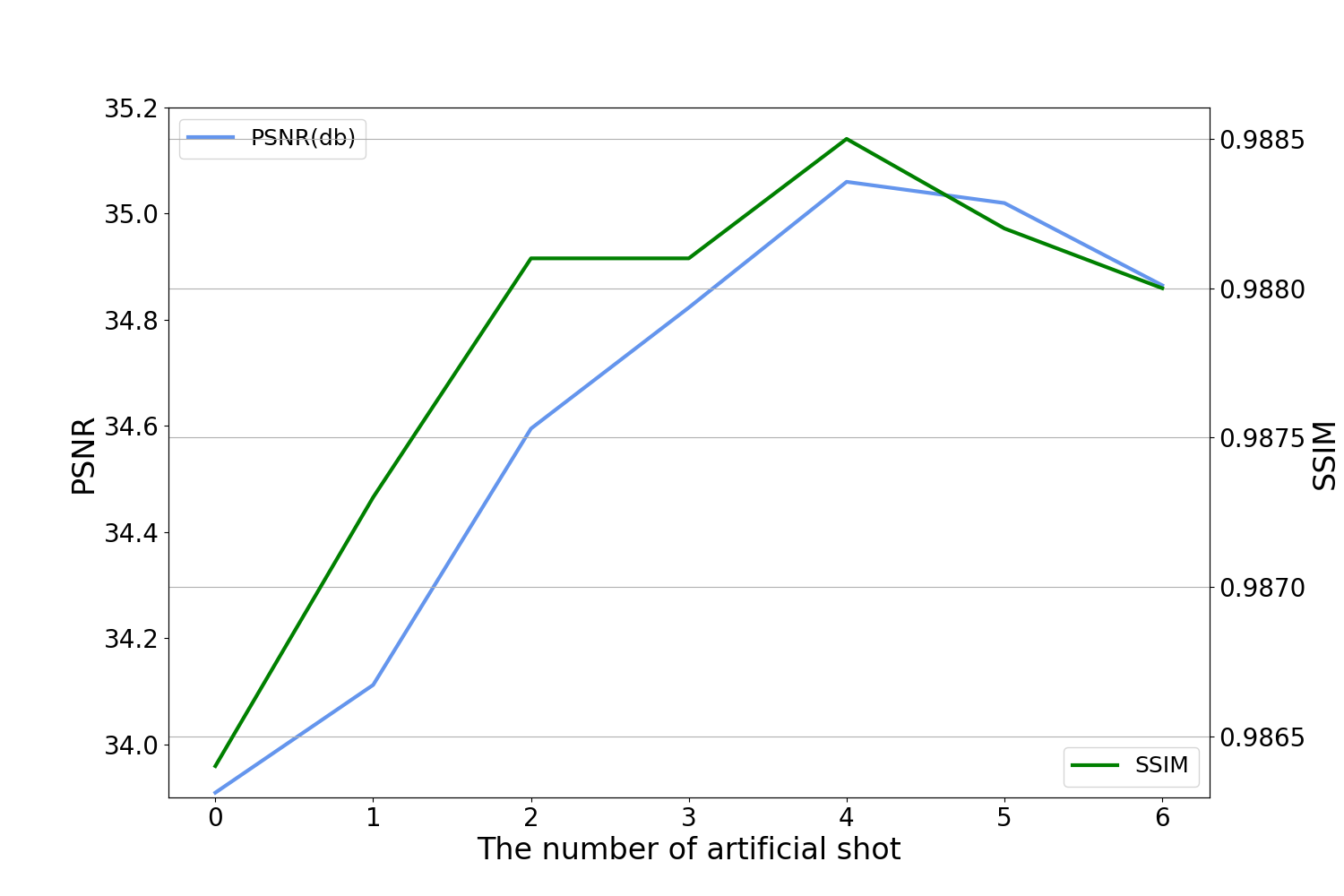}\caption{PSNR and SSIM performance on different numbers of artificial shot}
    \label{fig:12}
\end{figure}

\section{Discussion}
{In our experiments, we have verified the benefits of artificial multiple shots and bidimensional graph reasoning in removing non-homogeneous haze. These proposed two modules are cost efficient for coping with diverse images with significant content changes. However, when a video is fed to the dehazing system, the high correlation between neighboring frames would significantly increase the computational redundancy for our artificial shot generation and graph construction. In our future work, we would explore more flexible parameter reuse strategies for artificial shot and graph reasoning in the video dehazing task. Meanwhile, we will also attempt to extend the graph reasoning based non-local information propagation from the  spatial domain to the temporal domain in the video dehazing.}

\section{Conclusion}
{This paper proposes a novel Non-Homogeneous Haze Removal Network (NHRN). Instead of following the bivariate estimation based atmospheric scattering model, our NHRN considers the poor visibility of haze as inappropriate exposure and attempts to correct it via joint artificial multi-exposure fusion and non-local filtering, which are achieved by the proposed artificial scene prior and bidimensional graph reasoning modules. To the best of our knowledge, this is the first exploration to remove non-homogeneous haze via the graph reasoning based framework. By means of the enriched artificial scene priors and comprehensive contextual information utilization, the proposed method efficiently overcomes the error accumulation and haze residue issues when a single image is available. In comparison with the state-of-the-art methods, our NHRN achieves 2.46dB and 0.0022 gains in terms of PSNR and SSIM on the SOTS database.}

\section*{Acknowledgment}
This work was supported in part by National Science Foundation of China under Grant 61971095, Grant 61831005, Grant 61871087, Grant 62071086, and Sichuan Science and Technology Program under Grant 2021YFG0296.

\ifCLASSOPTIONcaptionsoff
  \newpage
\fi

\bibliographystyle{IEEEtran}
\bibliography{IEEEabrv, IEEEexample}

\end{document}